%% file: main.tex
\pdfoutput=1
\PassOptionsToPackage{table,dvipsnames}{xcolor}
\documentclass[10pt, logo, copyright]{nvidiatechreport}
\usepackage{url}
\usepackage{nicefrac}       %
\usepackage{multirow}
\usepackage{multicol}
\usepackage{xspace}
\usepackage{adjustbox}
\usepackage{tcolorbox}
\usepackage{wrapfig}
\usepackage{subcaption}    %
\usepackage{float}
\usepackage{stfloats}
\usepackage{listings}
\usepackage{arydshln}
\usepackage{color,soul}
\usepackage{makecell}
\usepackage[authoryear,sort&compress,round]{natbib}
\usepackage{algorithm}
\usepackage{algpseudocode}
\usepackage{placeins}
\usepackage{etoolbox}
\usepackage[capitalise,noabbrev]{cleveref}
\usepackage{refcount}

\makeatletter
\renewcommand{\maketitle}{\bgroup\setlength{\parindent}{0pt}
  \begin{adjustwidth}{0pt}{24pt}
    \begin{flushleft}
      {\raggedright \titlefont \bfseries \color{nvidiagreen} \@title\par}%
      \vskip11pt
      {\raggedright \bfseries \@author\par}
      \vskip8pt%
    \end{flushleft}
  \end{adjustwidth}
  \egroup
  \thispagestyle{firststyle}
}
\makeatother

\fancypagestyle{firststyle}{
  \fancyhf{}
  \fancyhead[L]{
    \ifthenelse{\boolean{logo}}{
      \includegraphics[width=90pt]{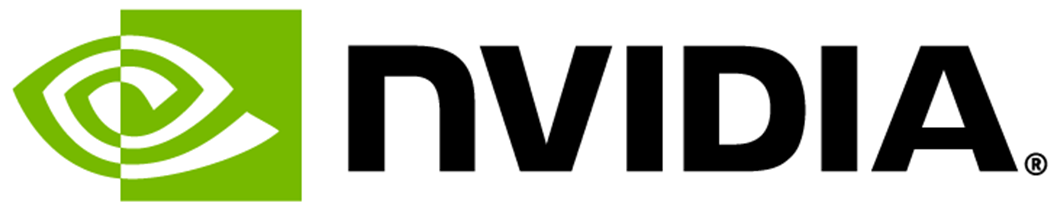}
    }{}
  }
  \fancyhead[R]{}
  \fancyhead[C]{}
  \fancyfoot[L]{
     $^{*}$Work done during an internship at NVIDIA \hspace{1em} \hfill \textcopyright\, \the\year{} NVIDIA. All rights reserved.
  }
  \fancyfoot[R]{}
  \fancyfoot[C]{}
}



\tcbuselibrary{listings,breakable}

\definecolor{pastelblue}{RGB}{173,216,230}
\definecolor{pastelyellow}{RGB}{255,253,208}
\definecolor{pastelpink}{RGB}{255,209,220}
\definecolor{pastelgreen}{RGB}{176,226,172}
\definecolor{pastellavender}{RGB}{230,230,250}
\definecolor{codebg}{RGB}{245,245,245}
\definecolor{codecomment}{RGB}{34,139,34}
\definecolor{codekeyword}{RGB}{0,80,160}
\definecolor{codestring}{RGB}{150,0,0}

\newcommand{\methodname}{DIRL}
\newcommand{\systemname}{Toolshed}
\newcommand{\modelname}{SpaceTools}

\newcommand{\gmark}{\textcolor{green!60!black}{\checkmark}}
\newcommand{\rmark}{\textcolor{red!70!black}{\tiny$\times$}}

\newcommand{\sy}[1]{}
\newcommand{\vb}[1]{}
\newcommand{\fl}[1]{}
\newcommand{\luke}[1]{}
\newcommand{\mika}[1]{}
\newcommand{\john}[1]{}

\title{\modelname: Tool-Augmented Spatial Reasoning via Double Interactive RL}

\author{Siyi Chen$^{1,2}$, ~~Mikaela Angelina Uy$^{1}$, ~~Chan Hee Song$^{1}$, ~~Faisal Ladhak$^{1}$, ~~Adithyavairavan Murali$^{1}$, ~~Qing Qu$^{2}$, ~~Stan Birchfield$^{1}$, ~~Valts Blukis$^{1}$$^{\dagger}$, ~~Jonathan Tremblay$^{1}$$^{\dagger}$ \\
\textsuperscript{1}NVIDIA \quad  \textsuperscript{2}University of Michigan \quad
$^{\dagger}$Project Leads
}

\begin{document}

\maketitle
\vspace{-3mm}
\noindent
\textbf{Links:}
{
\hypersetup{urlcolor=nvidiagreen}
\href{https://spacetools.github.io}{\textbf{Project Page}} |
\href{https://github.com/spacetools/SpaceTools}{\textbf{Code}}
}

\begin{figure}[h!]
    \centering
    \includegraphics[width=\linewidth]{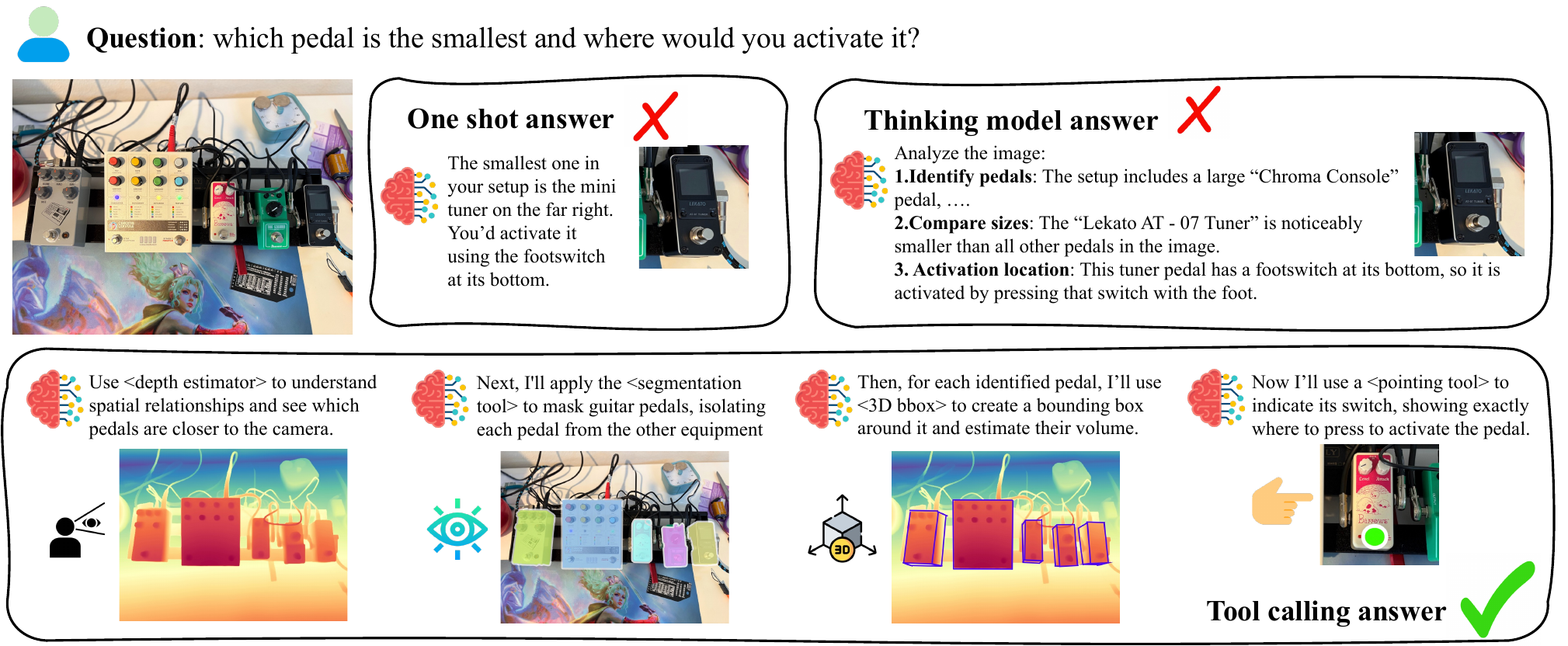}
    \caption{\modelname{} uses multiple computer vision tools to solve complex problems. We show a challenging example in which precise 3D understanding is required to identify the true smallest pedal.}
    \label{fig:teaser}
\end{figure}

\begin{abstract}
\textbf{Abstract:}
\textbf{Vision Language Models (VLMs) demonstrate strong qualitative visual understanding, but struggle with metrically precise spatial reasoning required for embodied applications.
The agentic paradigm promises that VLMs can use a wide variety of tools that could augment these capabilities, such as depth estimators, segmentation models, and pose estimators.
Yet it remains an open challenge how to realize this vision without solely relying on handcrafted prompting strategies or enforcing fixed, predefined tool pipelines that limit VLMs' ability to discover optimal tool-use patterns.
Reinforcement Learning could overcome this gap, but has so far been limited to reasoning with a single visual tool due to the large search space in multi-tool reasoning. 
We introduce Double Interactive Reinforcement Learning (DIRL), a two-phase training framework where VLMs learn to coordinate multiple tools through interactive exploration and feedback. In the teaching phase, we combine demonstrations from a single tool specialist trained via interactive RL with traces from a frontier model using all tools. 
In the exploration phase, the model further refines multi-tool coordination through continued RL. 
Our model, \modelname{}, with tool-augmented spatial reasoning ability, achieves state-of-the-art performance on spatial understanding benchmarks (RoboSpatial-Home, BLINK, BOP-ASK) and demonstrates reliable real-world manipulation using a 7-DOF robot as a tool. DIRL provides substantial improvements over the vanilla SFT (+12\% on RoboSpatial) and RL (+16\% on RoboSpatial) baselines.}
\end{abstract}
\abscontent

\input{tex/1_introduction}

\input{tex/2_related_work}
\input{tex/3_problem_formulation}

\input{tex/4_method}

\input{tex/5_experiments}
\input{tex/6_discussion}

\section*{Acknowledgments}
The authors would like to thank Vineet Bha, Alex Zook, Stephen Tyree, and Huijie Zhang for their inputs and comments on this manuscript.  

\bibliographystyle{plainnat}
\bibliography{paper}

\FloatBarrier
\clearpage

\appendix
\input{tex/appendix/0_main}

\end{document}

%% file: tex/1_introduction.tex
\section{Introduction}
\label{sec:intro}

Spatial reasoning—the ability to understand geometric relationships between objects and their environment—is an important capability for vision-language models (VLMs).
It enables models to answer geometric questions, such as relative positions, spatial configurations, and physical affordances, which is vital to support the integration of VLMs into embodied systems, such as robots.
While recent VLMs \cite{Lin2023VILA,liu2023improvedllava,bai2025qwen25vltechnicalreport,openai2025gpt5systemcard} have achieved strong performance on open-ended visual questions, their ability to do spatial understanding remains an active field of research \cite{kamath2023whatsup,OpenEQA2023,zhang2025from,yang2025embodiedbench}, particularly in settings that require diverse multi-step reasoning intertwined with precise geometric perception and 3D awareness (see Figure~\ref{fig:teaser}). 
These challenges are amplified in robotics, where perception must seamlessly translate into decision-making and physical action~\cite{kim2024openvla}.

The conventional approach to teach VLMs new capabilities involves fine-tuning on task-specific datasets~\cite{cheng2024spatialrgpt,ray2025sat,ji2025robobrain,zhou2025roborefer,cai2025depthlmmetricdepthvision,ssr}, an approach limited by the need for large-scale annotations and extensive data engineering.
We present a scalable alternative: we empower VLMs to use tools, that is, to call computer vision and robotics modules when needed, and use their outputs to aid in solving the spatial reasoning task.
Such tool use provides access to precise measurements and intermediate geometric representations, can leverage computer vision models from VLM-incompatible settings (\textit{e.g.}, dense prediction), and allows combining the strengths of heterogeneous models to augment base-model capability.
Recently, ViGoRL~\cite{ViGoRL_2025_arXiv} demonstrated that reinforcement learning can enable a VLM to learn grounded reasoning with a single visual tool, namely a cropping operation, showing the promise of interactive RL for tool use. 
However, naïve application of RL to many tools creates a prohibitively large search space where exploration fails to discover effective policies.

To address this gap, we introduce Double Interactive Reinforcement Learning (\methodname{}), a two-phase framework where interactive RL is applied twice.\textsuperscript{1} 
The key insight is that RL with a pointing tool is tractable and teaches grounding, 
while multi-tool RL can refine diverse reasoning, but requires good initialization for stable learning. 
\methodname{} uses a two-phase training scheme with a teaching phase followed by an exploration phase.
In the teaching phase, the model is trained with Supervised Fine-Tuning (SFT) on the basics of tool usage—method signatures, outputs, and information flow using a mix of single-tool Interactive Reinforcement Learning (IRL) traces and multi-tool demonstrations.
In the exploration phase, we apply Interactive Reinforcement Learning (IRL) with the full toolset, enabling the model to refine tool coordination for spatial reasoning tasks.

Unlike prior work, \methodname{} allows the model to call tools interactively during training, instead of relying on fixed pipelines or precomputed contexts (Table~\ref{tab:tool_calling_comparison}), 
enabling this behavior at scale requires addressing a key systems challenge: how to efficiently serve diverse, compute-intensive tools during interactive training.
To address this, we develop  
\textit{\systemname{}}, a platform which hosts computationally intensive computer vision tools such as SAM2~\cite{ravi2024sam2}, Depth Pro~\cite{depthpro2025}, RoboRefer~\cite{zhou2025roborefer}, and GraspGen~\cite{graspgen} as rapid on-demand services during training, decoupling tool resource management from RL or inference workloads, and achieving high tool throughput and utilization.
By incorporating real and stochastic tool outputs into the learning loop, \methodname{} exposes models to actual tool behavior, encouraging reasoning about tool reliability and discovering improved ways to query the tools.

We conduct extensive experiments on a diverse set of spatial reasoning problems, such as determining object-location fit, estimating distances between items, reasoning about occlusions and orientations, pose estimation, and predicting grasping affordances.
Our trained model, \textit{\modelname}, achieves state-of-the-art performance across multiple spatial reasoning benchmarks, including RoboSpatial-Home~\cite{song2025robospatial}, BLINK~\cite{fu2024blink}, RefSpatial~\cite{zhou2025roborefer}, CVBench~\cite{zhou2025roborefer}, 
and BOP-ASK~\cite{bhat2025bopask}. 
By integrating a real robot as a tool, \modelname{} completes pick-and-place tasks with an 86\% success rate, demonstrating effective transfer from spatial reasoning to embodied control and outperforming frontier models equipped with the same tools.
In summary, our contributions are: 
\begin{enumerate}
    \item \textbf{\methodname:} a novel training paradigm that enables interactive training with a large set of tools. 
    \item \textbf{\systemname:} an interactive platform for hosting diverse computer vision tools, to be open-sourced.
    \item \textbf{\modelname:} A VLM trained for spatial reasoning via interactive multi-tool use, which achieves state-of-the-art results across spatial reasoning benchmarks and performs robot control via alternating perception and action tool calls. 
\end{enumerate}

%% file: tex/2_related_work.tex
\begin{table}[h!]
\centering
\footnotesize 

\caption{Comparison of related work for training supervision and tool-call interactivity during training. `-' indicates that only a single tool is used.  
}
\begin{tabular}{lccccc}
\toprule
\textbf{Method} &
\textbf{SFT} &
\textbf{RL} &
\makecell{\textbf{Use}\\\textbf{tools}} &
\makecell{\textbf{Non-fixed}\\\textbf{tool pipeline}} &
\makecell{\textbf{Interactive}\\\textbf{tool call}} \\
\midrule
SpatialVLM~\cite{chen2024spatialvlm}        & \gmark & \rmark & \rmark & \rmark & \rmark \\
RoboRefer~\cite{zhou2025roborefer}          & \gmark & \gmark & \rmark & \rmark & \rmark \\ 
SpatialPIN~\cite{ma2024spatialpin}          & \rmark & \rmark & \gmark & \rmark & \rmark \\
APC~\cite{lee2025apc}                       & \rmark & \rmark & \gmark & \rmark & \rmark \\
ViGoRL~\cite{ViGoRL_2025_arXiv}               & \rmark & \gmark & - & \gmark & \rmark \\
SpatialReasoner~\cite{ma2025spatialreasoner} & \rmark & \gmark & \rmark & \rmark & \rmark \\
\modelname{} (ours)                        & \gmark & \gmark & \gmark & \gmark & \gmark \\
\bottomrule
\end{tabular}
\label{tab:tool_calling_comparison}
\end{table}

\section{Related Work}
\label{sec:related_work}

\paragraph{Spatial Reasoning with VLMs.} Spatial reasoning with VLMs \cite{Liu2024LLavaNeXT,Lin2023VILA,liu2023improvedllava,bai2025qwen25vltechnicalreport,openai2025gpt5systemcard} refers to understanding geometric relationships among objects and their environment \cite{fu2024blink,scanqa,szymanska2024space3dbench,ramakrishnan2025space,empirical,li2025imaginereasoningspacemultimodal,mindseye}. 
Recent progress shows that VLMs can increasingly support robots in perceiving and interacting with the physical world \cite{spatialbot,song2025robospatial,zhou2025roborefer}. However, VLM spatial reasoning remains insufficient for real-world robotic demands, where multi-step reasoning, precise geometric understanding, and strong 3D awareness are required \cite{CosmosReason1,zhang2025from,yang2025embodiedbench}. Conventional approaches teach VLMs spatial understanding by fine-tuning on task-specific question-answering datasets \cite{Chen_2024_CVPR,cheng2024spatialrgpt,song2025robospatial,ray2025sat,wu2025reinforcingspatialreasoningvisionlanguage,zhou2025roborefer,ViGoRL_2025_arXiv,liu2025spatialcot}. Yet these methods require large-scale data collection and architecture modifications even to introduce a single low-level perceptual capability such as depth \cite{cai2025depthlmmetricdepthvision}, pointing \cite{song2025robospatial,zhou2025roborefer,Deitke2025Molmo}, and 3D-awareness \cite{SpatialLLM_2025_arXiv,Ma_2025_SpatialReasoner_arXiv}. Instead of baking all perceptual skills into the model, we propose to enable VLMs to invoke external computer vision and robotics tools as needed, allowing them to solve spatial reasoning tasks and perform real-world manipulation.

\paragraph{Tool-augmented Reasoning.}
Tool-augmented reasoning aims to enrich model capabilities by supplying additional information from external modules \cite{liu2024toolace,zhang2024xlam,research,jin2025search}.
Typical applications include integrating search engines \cite{retool,research,jin2025search}, calculators \cite{zhang2025nemotronresearchtooln1exploringtoolusinglanguage,webgpt}, or code executors \cite{suris2023vipergpt,wang2024executable} into LLMs, and vision tools for VLMs \cite{hu2024visual,ToolVQA_2025_ICCV,ma2024mmsbenchmarkevaluatetooluse}.
In the context of spatial reasoning, the community has explored equipping VLMs with vision tools during intermediate reasoning steps. 
However, most approaches rely on 
handcrafted prompting strategies \cite{yang2023setofmarkpromptingunleashesextraordinary,hu2024visual,marsili2025visual,visprog}
or enforce a fixed, predefined tool pipeline \cite{ma2024spatialpin,lee2025apc} in a training-free way,
which limits their ability to handle diverse, precise, and 3D-aware reasoning required for robotics. 
In contrast, we enable the model to learn to coordinate a diverse set of vision and robotic tools through both teacher demonstrations involving real tool interactions and self-exploration enabled by interactive RL. We also acknowledge concurrent and subsequent works that explore complementary directions: TIGeR~\cite{han2025tiger} studies tool-augmented reasoning learned from synthetic tool-use traces, Think3D \cite{zhang2026think3dthinkingspacespatial} focuses on reasoning with 3D manipulation tools, and Cap-X \cite{fu2026capxframeworkbenchmarkingimproving} emphasizes benchmarking code-generation-based tool use for robotic manipulation.

\paragraph{Reinforcement Learning for Reasoning.}
Reinforcement learning (RL) has been widely applied to enhance the reasoning capabilities of LLMs or VLMs on verifiable tasks such as math \cite{grpo}, coding \cite{deepseekr1,kimik1_5}, and general visual question answering (VQA) \cite{yang2025r1onevision,Zhang2025R1VL,vision-r1,Zhang2025ChainofFocusAV}. Recent work further explores RL for spatial reasoning, enabling models to produce interpretable or grounded reasoning \cite{liu2025visual,vlm-r1,wu2025reinforcingspatialreasoningvisionlanguage,vtool-r1,kim2025robotr}.
Some works adopt RL to stengthen chain-of-thought style reasoning before predicting answers \cite{Wang_2025_SVQA-R1_arXiv,CosmosReason1}, while others focus on teaching grounded spatial understanding \cite{ViGoRL_2025_arXiv,wu2025reinforcingspatialreasoningvisionlanguage,deepeyes}. 
Although prior works demonstrate that RL can teach spatial reasoning with use of a single light-weight tool (\textit{e.g.}, cropping), 
scaling to multiple heterogeneous tools poses a fundamental challenge: with 10+ tools, the action space grows combinatorially, causing naive RL exploration to fail. 
Our training paradigm decomposes the problem into progressive and tractable phases, enabling the model to learn effective coordination strategies with diverse tools.

%% file: tex/3_problem_formulation.tex
\begin{algorithm}[t]
\small
\caption{Spatial Reasoning with Tools}
\begin{algorithmic}[1]
\Require VLM $\pi_{\theta}$, User Query $\mathcal{I}$, Max Turns $T_{\max}$
\Ensure Answer $A_{\text{final}}$
\State $t \gets 1$,  $h_1 \gets \mathcal{I}$ \Comment{Initialize dialogue history}
\While{$t \leq T_{\max}$} \Comment{\textit{t} is a counter}
  \State $a_t \gets \pi_{\theta}(h_{t})$ \Comment{Generate VLM response}
  \State $h_{t+1} \gets h_{t} \oplus a_t$
  \If{\textcolor{RoyalBlue}{\texttt{<answer>}} detected in $a_t$}
    \State $A_{\text{final}} \gets \text{Parse}(a_t, \textcolor{RoyalBlue}{\texttt{<answer>}},\textcolor{RoyalBlue}{\texttt{</answer>}})$
    \State \textbf{break} \Comment{Final turn: task is complete}
  \ElsIf{\textcolor{LimeGreen}{\texttt{<tool\_call>}} detected in $a_t$}
    \State $\mathcal{Q}_{\text{tools}} \gets \text{Parse}(a_t, \textcolor{LimeGreen}{\texttt{<tool\_call>}},\textcolor{LimeGreen}{\texttt{</tool\_call>}})$
    \For{\textbf{each} $q \in \mathcal{Q}_{\text{tools}}$}
      \State $h_{t+1} \gets h_{t+1} \oplus \text{CallTool}(q)$
      \Comment{Execute tool}
    \EndFor
  \EndIf
  \State $t \gets t+1$
\EndWhile
\State \Return $A_{\text{final}}$
\end{algorithmic}
\label{alg:iterative_tool}
\end{algorithm}

\section{Problem Formulation}
\label{sec:problem_formulation}

We formulate spatial reasoning as a sequential decision-making problem where a VLM policy $\pi_{\theta}$ interacts with external tools $\mathcal{Q}_\textit{tools}$ to respond to a user query $\mathcal{I}$, which may consist of an image-text pair or a robotic manipulation task. The model can reason and interact with tools in multiple turns until it produces a final answer $A_{\text{final}}$ or reaches a maximum of $T_{\max}$ interaction steps.

At each step $t$, the VLM receives the historical context $h_t$, which contains the full dialogue between the user, the VLM, and the tools (initialized as $h_1 = \mathcal{I}$). The model then generates a response $a_t$ according to its policy: 
If $a_t$ includes tool calls, tools are executed sequentially. Their outputs, together with $a_t$, are appended to the historical context $h_t$ to form $h_{t+1}$. The updated context is then used to generate the next-step response.

The complete workflow is outlined in Algorithm~\ref{alg:iterative_tool}. The model is required to follow a structured conversational format: reasoning is enclosed within \textcolor{Orange}{\texttt{<think>}} tags, tool calls within \textcolor{LimeGreen}{\texttt{<tool\_call>}} tags, and the final answer within \textcolor{RoyalBlue}{\texttt{<answer>}} tags. The goal of this work is to \textbf{learn a policy} $\pi_{\theta}$ that addresses user
queries through multi-turn interaction with vision and robotic tools. To achieve this, we propose a new training paradigm
accompanied by a novel tool platform.

%% file: tex/4_method.tex
\section{Double Interactive Reinforcement Learning}
\label{sec:method}

Training a VLM to reason and act through external tools benefits from both teacher-guided supervision and interactive exploration. We introduce \textbf{Double Interactive Reinforcement Learning (\methodname{})}, a two-stage framework that unifies these two forms of learning. 
Enabling \methodname{} requires seamless communication between the VLM and a diverse set of vision and robotic tools during both data collection and training. 
We solve this challenging problem by designing \textbf{\systemname{}}, a distributed infrastructure that manages large-scale tool interaction.

\subsection{\methodname{}}

We introduce a new training paradigm that enables VLMs to effectively use multiple tools. 
Our approach is motivated by two observations: 
(i) naïvely applying IRL (interactive RL) to all tools at once creates an extremely large search space, resulting in weak optimization signals, and 
(ii) pure SFT on tool-interaction traces yields models that struggle to coordinate with tools effectively or to go beyond the training traces. 
Our method, \methodname{}, addresses these limitations and improves the model's ability to integrate and sequence multiple tools effectively.
\methodname{} is composed of two phases, a teaching phase and an exploration phase.

\paragraph{Teaching phase.}
This phase establishes basic tool use capabilities without the exploration challenges of full multi-tool RL. We build the teaching dataset from two complementary sources.
First, we apply IRL to train the base model to use a single pointing tool for spatial reasoning tasks (\textit{e.g.}, spatial relationship, spatial compatibility, and relative depth are trained together).
This constrained search space, allows IRL to reliably converge and produce competent behavior. 
The resulting \textit{IRL-trained teacher} is then used to generate supervised demonstrations of grounded reasoning for the first portion of our teaching dataset.
Second, we prompt a \textit{universal teacher}, which is a frontier model, to solve spatial reasoning and robot manipulation tasks with the full set of tools (\textit{e.g.}, pointing, segmentation, 3D bbox, \textit{etc.}), 
retaining only trajectories that lead to correct solutions.
Finally, we combine both datasets—one part generated by the IRL-trained teacher and three parts from the universal teacher—to form the complete teaching dataset. We then perform supervised fine-tuning (SFT) on the base model, yielding a policy with initial tool-usage behaviors.

\paragraph{Exploration phase.}
This phase refines multi-tool coordination through interactive exploration. 
We resume IRL training on all tasks from the SFT-initialized policy 
with access to all available tools, allowing the model to enhance tool chaining strategies. The strong initialization prevents exploration collapse in the large multi-tool action space, while interactive feedback offers additional refinement of tool coordinations.
These two rounds of IRL give our method its name, as \methodname{} involves two IRL phases—one for teaching and one for exploration.

\begin{figure}[h!]
    \centering
    \includegraphics[width=0.6\linewidth]{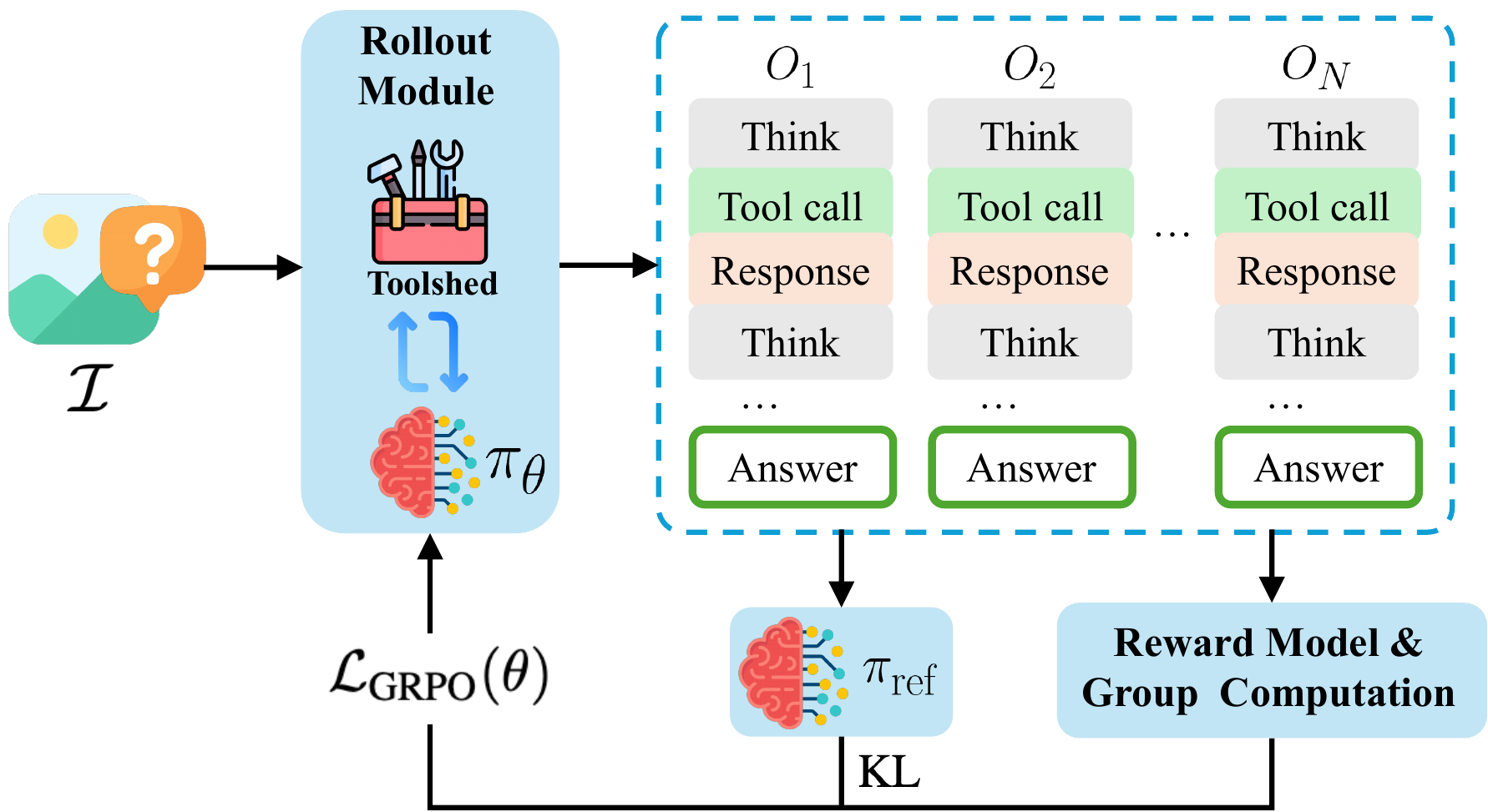}
    \caption{Interactive reinforcement learning (IRL) with \systemname{}.
    The rollout module executes multi-turn trajectories under policy $\pi_{\theta}$, alternating between reasoning and tool use before answering. Task rewards are aggregated and used to update $\pi_{\theta}$ via GRPO with KL regularization against $\pi_{\text{ref}}$.
    }
    \label{fig:rl_training}
\end{figure}

\paragraph{Policy Update.}
We employ Group Relative Policy Optimization (GRPO)~\cite{grpo} as our RL training algorithm, as visualized in \Cref{fig:rl_training}.
For each input $\mathcal{I}$, in total $N$ rollout procedures are launched asynchronously under the current policy $\pi_{\theta}$. Each rollout proceeds as \Cref{alg:iterative_tool}, generating in total $N$ multi-turn rollouts $O_1$, $O_2$, $\dots$, $O_N$. Their rewards are calculated as $r_1$, $r_2$, $\dots$, $r_N$, and the policy is updated by optimizing the GRPO objective
$\mathcal{L}_{\text{GRPO}}(r_1, \dots, r_N)$, described in full in 
the Appendix.

\subsection{\systemname{}}
Our method, \methodname{}, assumes access to an efficient system for invoking tools during training.
In prior work, tool usage is either tightly coupled with the training loop, limiting to simple tools (\textit{e.g.}, cropping~\cite{ViGoRL_2025_arXiv}), or in case of text-only tools (\textit{e.g.}, search~\cite{jin2025search}), highly decoupled via web APIs that lack the throughput needed for VLM interactive learning with images.
Others side-step the issue altogether by using pre-computed tool outputs \cite{han2025tiger}, preventing models from learning interactive, state-dependent tool use.

We introduce \systemname{}, 
a scalable framework for deploying multiple compute-heavy tools alongside training or inference workloads
that mitigates these bottlenecks through:
(1) resource and environment isolation for each tool instance;
(2) decoupled scaling and execution from the policy's main inference loop; and
(3) asynchronous parallel workers per tool, allowing scaling tool resources independently from training resources.
\systemname{} hosts modular vision tools (\textit{e.g.}, segmentation, pointing, monocular depth, 3D box fitting, grasp prediction, and various image operations) and robotic tools (\textit{e.g.}, image capture, grasp execution, object placement).
Implementation details and complete tool APIs are provided in 
the Appendix.

\input{Tables/main_table}

\subsection{Rewards}
Reinforcement learning covers spatial reasoning tasks such as multiple choice question answering, 2D bounding box localization, pointing, pose, and grasp estimation. We design normalized, task-specific rewards based on the correctness of the final answer $A_{\text{final}}$.
Each reward measures the accuracy or geometric consistency of $A_{\text{final}}$ against the ground-truth label or annotation. We additionally experimented with a structural \emph{format score} to encourage output correctness, but found it provided no measurable improvement and excluded it from final training. Details in the Appendix.

\begin{itemize}[leftmargin=*]

\item \textbf{Multiple choice questions.}
The reward is binary: 
$R_{\text{B}} = 1$ if $A_{\text{Final}}$ is correct, else $0$.

\item \textbf{2D bounding boxes.}
We compute Mean IoU (MIoU) between predicted and ground-truth boxes:
$R_{\text{MIoU}} = \frac{1}{N}\sum_{i=1}^{N} \max_j \mathrm{IoU}(\hat{B}_i, B_j)$,
where $\hat{B}_i$ and $B_j$ denote predicted and ground-truth boxes.

\item \textbf{Pointing.}
For single-point spatial prediction, we use the Normalized Negative Distance to Centroid (NNDC):
$R_{\text{NNDC}} = \frac{\exp(-5d) - \exp(-5\sqrt{2})}{1 - \exp(-5\sqrt{2})}$,
where $d$ is the distance to the target-region centroid.
To emphasize precision, we clip with the binary accuracy term:
$R = \max(R_{\text{NNDC}}, R_{\text{B}})$.

\item \textbf{Pose estimation.}
Predicted and ground-truth poses are converted to eight 2D projected corners.
The reward is the IoU between convex hulls of predicted ($\hat{C}$) and ground truth ($C$) corner sets.
$R_{\text{IoU}} = \mathrm{IoU}(\hat{C}, C)$
when both sets are valid ($|\hat{C}| = |C| = 8$), and $0$ otherwise.

\item \textbf{Grasp estimation.}
We adopt the Normalized Negative Coordinate Error (NNCE):
$$R_{\text{NNCE}} = 1 - \frac{1}{\delta_{\max}}
\min\!\left(\delta_{\max}, \frac{1}{N}\sum_{i=1}^{N}
\frac{\lVert \hat{p}_i - p_i \rVert_2}{d}\right)$$,
where $\hat{p}_i$ and $p_i$ are predicted and ground-truth contact points,
$d$ is the gripper width, $N$ is the number of reference points,
and $\delta_{\max}$ caps extreme errors.
This rewards accurate geometric grasp alignment.
In this work, $\delta_{\max}=10$.

\end{itemize}

%% file: Tables/main_table.tex
\begin{table}[h!]
\centering
\scriptsize
\caption{Performance comparison across spatial reasoning benchmarks. All values are normalized accuracy (\%). \textbf{Bold} indicates the best performance within each column, and \underline{underline} denotes the second-best result. Values of 0 indicate the model either fails to produce valid responses, outputs answers in wrong formats, or produces entirely incorrect predictions, reflecting an inability to handle that task type.
}
\resizebox{\textwidth}{!}{
\begin{tabular}{@{}lcccccccccc@{}}
\toprule
\multirow{2}{*}{\textbf{Model}} 
& \multicolumn{3}{c}{\textbf{RoboSpatial}}  
& \textbf{BLINK}
& \multirow{2}{*}{\textbf{RefSpatial}} 
& \multicolumn{2}{c}{\textbf{CVBench}} 
& \multicolumn{3}{c}{\textbf{BOP-ASK}}  \\ 
\cmidrule(lr){2-4} \cmidrule(lr){5-5} \cmidrule(lr){7-8} \cmidrule(lr){9-11}
 & \textbf{VQA} & \textbf{Vacant} & \textbf{Overall} 
 & \textbf{Depth} 
 &  & \textbf{2D Rel. } & \textbf{3D Depth } & \textbf{Pose} & \textbf{Grasp-MACE} & \textbf{Grasp-SR} \\ 
\midrule

\rowcolor[HTML]{F5F5F5}
\multicolumn{11}{c}{\textit{Proprietary Models}} \\
Claude Sonnet 4.5 & 75.44 & 23.77 & 57.43 & 78.23 & \phantom{0}7.49 & 89.85 & 78.50 & \phantom{0}1.67 & 40.12 & \underline{48.33} \\
GPT-4o & 61.61 & 25.10 & 48.88 & 63.71 &  \phantom{0}8.48 & 88.77 & 75.50 & \phantom{0}0.00 & \phantom{0}5.50 & \phantom{0}1.67 \\
GPT-5 & 76.50 & 22.17 & 58.39 & 66.13 & 23.10 & \underline{95.54} & 91.33 & \phantom{0}9.03 & \underline{39.59} & 41.67 \\
Gemini-ER 1.5 & \underline{79.30} & 31.10 & \underline{62.50} & 69.23 & 41.72 & \underline{95.54} & 90.50 & \phantom{0}0.00 & 30.06 & 23.33  \\[2pt]
\midrule
\rowcolor[HTML]{F5F5F5}
\multicolumn{11}{c}{\textit{General Open-Source Models}} \\ 
LLaVA-NeXT-8B & 69.31 & \phantom{0}0.00 & 45.15 & 53.23 & \phantom{0}0.78 & 72.15 & 73.67 & \phantom{0}0.00 & \phantom{0}5.04 & \phantom{0}1.67 \\
Qwen2.5-VL-32B & 61.84 & \phantom{0}3.28 & 41.43 & 70.16 & \phantom{0}7.28 & 90.46 & 86.67 & \phantom{0}0.00 & 29.86 & 23.33 \\
Qwen2.5-VL-3B & 53.07 & \phantom{0}0.00 & 35.71 & 70.98 & \phantom{0}0.00 & 70.62 & 65.33 & \phantom{0}0.00 & \phantom{0}6.06 & \phantom{0}0.00  \\
[2pt]
\midrule
\rowcolor[HTML]{F5F5F5}
\multicolumn{11}{c}{\textit{Spatial VLMs}} \\ 
SpaceLLaVA-13B & 61.00 & \phantom{0}2.50 & 40.61 & 51.61 & \phantom{0}3.25 & 61.08 & 62.83 & \phantom{0}0.00 & \phantom{0}0.00 & \phantom{0}0.00 \\
RoboPoint-13B & 70.18 & 19.70 & 52.58 & 54.84 & 15.59 & 74.00 & 76.50 & \phantom{0}0.00 & \phantom{0}0.00 & \phantom{0}0.00 \\
Molmo-7B & 39.92 & \phantom{0}0.82 & 26.29 & 54.03 & \phantom{0}0.00 & 72.15 & 73.33 & \phantom{0}0.00 & 36.74 & 18.33 \\
RoboBrain2.0-7B & 59.64 & 44.35 & 54.31 & 84.68 & 32.50 & 87.23  & 90.00 & \phantom{0}0.00 & \phantom{0}0.00 & \phantom{0}0.00 \\
RoboRefer-8B-SFT & 58.33 & \textbf{61.48} & 59.43 & \underline{88.71}  & \underline{48.37} & \textbf{96.31} & \textbf{96.50} & \phantom{0}0.00 & \phantom{0}0.00 & \phantom{0}0.00 \\
[2pt]
\midrule
\rowcolor[HTML]{F5F5F5}
\multicolumn{11}{c}{\textit{Tool-free Fine-tuning}} \\ 

Qwen2.5-VL-3B-Tool-free SFT & 66.66 & 41.80 & 58.00 & 80.65 & 20.22 & 91.54 & 83.33 & \phantom{0}2.44 & 39.47  & 35.00 \\
Qwen2.5-VL-3B-Tool-free RL & 67.54 & 28.69 & 54.00 & 80.65 & 23.10 & 87.38 & 70.83 & \underline{12.00} & 38.79 & 36.67  \\
[2pt]


\midrule
\textbf{\modelname-3B (Ours)} & \textbf{79.38} & \underline{52.46} & \textbf{70.00} & \textbf{90.32} &  \textbf{53.07} & 94.92 & \underline{96.00} & \textbf{34.37} & \textbf{43.06} & \textbf{50.00} \\

\bottomrule
\end{tabular}}
\label{tab:spatial_vlm_comparison}
\vspace{-.1in}
\end{table}

%% file: tex/5_experiments.tex
\section{Experiments}
\label{sec:experiments}

\paragraph{Dataset.}
During the first phase of \methodname{}, we generate a teaching SFT dataset composed of 8k high-quality tool-use trajectories: 6k from the universal teacher and 2k from the IRL-trained teacher. The IRL teacher is trained to use a Pointing tool (RoboRefer~\cite{zhou2025roborefer}), a common first step before querying other vision and robotic tools in spatial reasoning. For the universal teacher, we use Claude Sonnet 4.5~\cite{anthropic2025claudesonnet4_5_systemcard}, integrated with \systemname{}, consisting of all tools.
Image–question pairs are sampled from RoboSpatial~\cite{song2025robospatial}, RefSpatial~\cite{zhou2025roborefer}, 
and BOP-ASK~\cite{bhat2025bopask}.
To extend our setup to robot manipulation, we augment the HOPE dataset~\cite{tyree2022hope} with grasping and pick-and-place control tasks. 
Each trajectory in the SFT dataset is a multi-turn dialogue between the user, assistant, and tools.
The same spatial reasoning image-question pairs in the teaching SFT dataset are used in the stage-2 IRL.

\begin{figure}[h!]
    \centering
    \includegraphics[width=\linewidth]{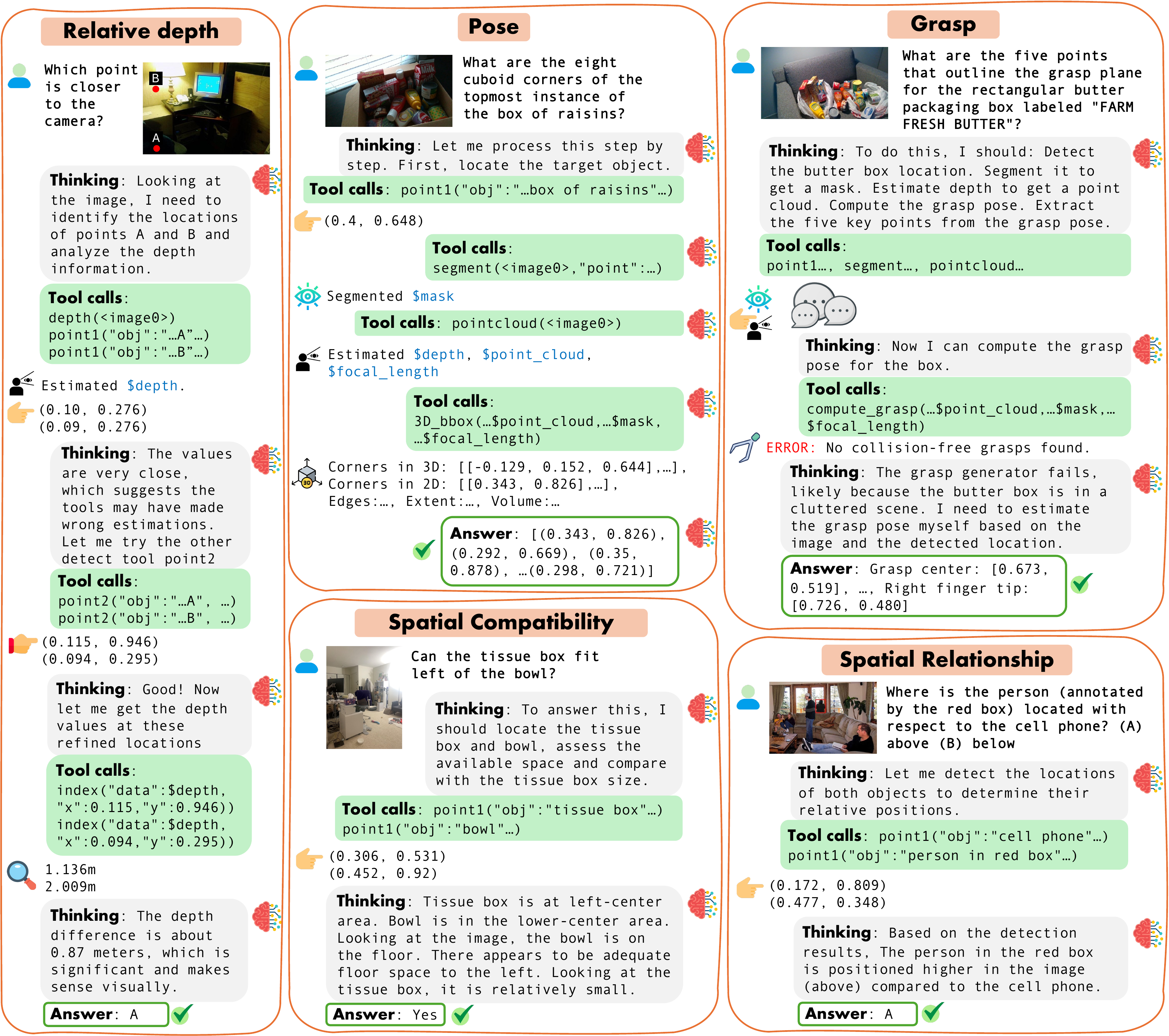}
    \caption{
        Spatial reasoning examples of \modelname{}. It performs diverse spatial reasoning tasks including relative depth, pose, grasp, spatial compatibility, and spatial relationship by interleaving reasoning (gray) and vision tool calls (green) before producing the final answer. Images are taken from BLINK \cite{fu2024blink}, RoboSpatial-Home \cite{song2025robospatial}, and 
        BOP-ASK~\cite{bhat2025bopask}.
    }
    \label{fig:traces}
\end{figure}

\paragraph{Training.}
We use Qwen2.5-VL-3B-Instruct \cite{bai2025qwen25vltechnicalreport} as our  base model. During training, we use a system prompt that defines the available tools, their schemas, and usage formats (detailed prompt is in the Appendix).
High-level instructions within the prompt further specify the model's reasoning structure and output format.
We use the following perception tools: segmentation~\cite{ravi2024sam2}, pointing~\cite{Deitke2025Molmo,zhou2025roborefer}, depth estimation~\cite{depthpro2025}, cuboid fitting, grasp generation~\cite{graspgen}, cropping, array indexing, perspective projection, and robot tools: image capture, depth capture, grasp, place.
The first-stage SFT starts with the base model and uses cross-entropy next-token prediction loss over all assistant turns in a multi-turn dialogue, training the policy $\pi_{\theta}$ to imitate the teachers' reasoning processes and tool-use behaviors.
In the second-stage IRL, the SFT-initialized model interacts with \systemname{} to refine its tool-augmented spatial reasoning through interactive feedback from task rewards.
For SFT, we use the LLaMA-Factory platform~\cite{zheng-etal-2024-llamafactory}, and for RL we integrate Toolshed into the VERL framework~\cite{sheng2024hybridflow}.
More implementation details are in 
the Appendix.

\begin{figure}[h!]
    \centering
    \includegraphics[width=\linewidth]{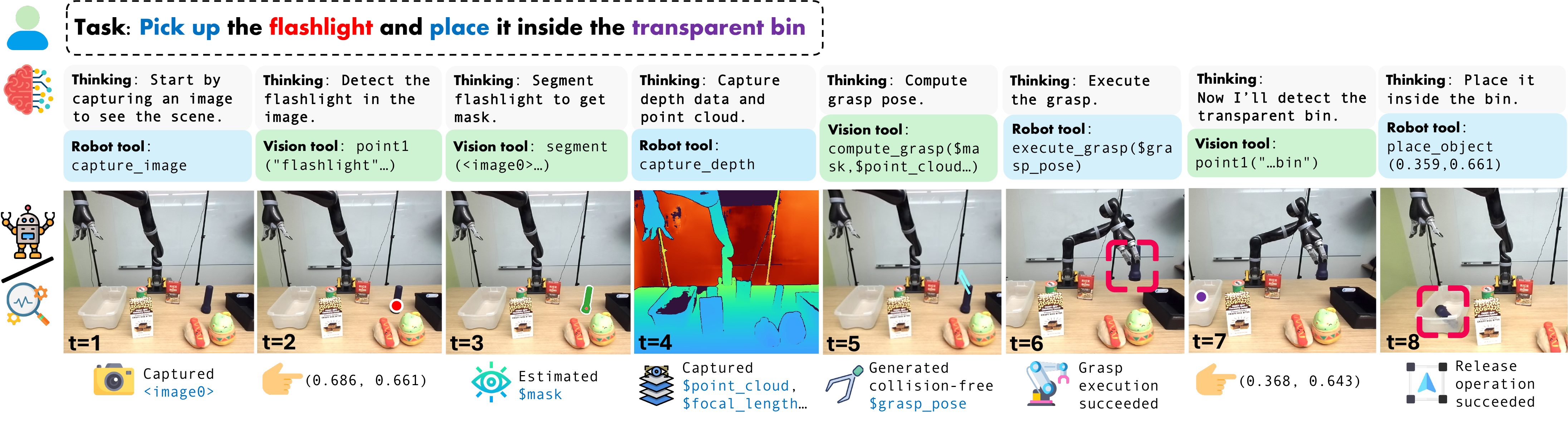}
    \caption{Real-world robot manipulation fully controlled by \modelname{}. The model completes a multi-step task, ``picking up the flashlight and placing it in the transparent bin'', via alternating reasoning (gray), vision tools (green) for perception, and robot tools (blue) for action.
    }
    \label{fig:robot_exp}
\end{figure}

\subsection{Spatial Reasoning Evaluation}

\paragraph{Benchmarks and Metrics.}
We evaluate our model on a suite of spatial reasoning benchmarks, including RoboSpatial-Home~\cite{song2025robospatial} (spatial VQA and vacant space pointing), CVBench~\cite{cvbench2025} (2D relations and 3D relative depth), RefSpatial~\cite{zhou2025roborefer} (placement, location, and unseen), BLINK~\cite{fu2024blink} (relative depth), 
and BOP-ASK~\cite{bhat2025bopask}.
They cover positional relationship understanding, depth estimation, pointing, 3D pose estimation, and robotic grasp prediction.
We adopt the following metrics:
(1) Answer accuracy for multiple-choice and pointing questions. 
(2) For object pose estimation, we use the normalized Intersection-over-Union (IoU) in  range $[0,100]$ (\%). 
(3) For grasp estimation, which outputs five 2D coordinates of grasp center, and two finger bases and tips, we use the \emph{Mean Angular Coordinate Error (MACE)} to jointly score grasp location and finger-orientation, defined in 
the Appendix. 
We report MACE as a normalized score in range $[0,100]$ (\%), and the \emph{Success Rate (SR)} as the percentage of grasps achieving $\text{MACE} > 40$.

\paragraph{Baselines.}
We compare our model (\modelname{}) against four categories of baselines. 
(1) \textbf{Proprietary models} include Claude Sonnet~4.5 \cite{anthropic2025claudesonnet4_5_systemcard}, GPT-4o \cite{openai2024gpt4techreport}, GPT-5 \cite{openai2025gpt5systemcard}, and Gemini-ER~1.5 \cite{geminiroboticsteam2025geminirobotics15pushing}, which represent state-of-the-art commercial vision-language systems. 
(2) \textbf{General open-source models} include LLaVA-NeXT-8B \cite{liu2023improvedllava} and Qwen2.5-VL-32B \cite{bai2025qwen25vltechnicalreport}, which serve as publicly available multimodal foundations without spatial specialization. 
(3) \textbf{Spatial VLMs} include SpaceLLaVA-13B \cite{chen2024spatialvlm}, RoboPoint-13B \cite{Yuan2024RoboPoint}, Molmo-7B \cite{Deitke2025Molmo}, RoboBrain2.0-7B \cite{robobrain2}, and RoboRefer-8B-SFT, which are trained with additional spatial reasoning or robotic data. 
(4) \textbf{Tool-free fine-tuning} contains variants of the same base model (Qwen2.5-VL-3B) trained without tool use, only on the 8k source question and answer samples\textsuperscript{2}\label{fn:answersamples} from DIRL's stage-1:
(4a) \textit{Tool-free SFT} is a supervised fine-tuning baseline. 
(4b) \textit{Tool-free RL} applies reasoning RL à la GRPO~\cite{deepseekr1} without tool use.

\subsection{Spatial Reasoning Results}
As shown in \Cref{tab:spatial_vlm_comparison}, \modelname{} achieves state-of-the-art results on nearly all benchmarks, surpassing proprietary, open-source, and  spatial VLM baselines.
\modelname{} outperforms Gemini-ER~1.5 by +7.5\% on RoboSpatial, exceeds Claude Sonnet~4.5 by +24.4\% on pose estimation, and surpasses GPT-5 by +8.3\% on grasp prediction.
Moreover, tool-augmented training yields substantially stronger results on spatial reasoning than tool-free fine-tuning of the same base model on the the same 8k VQA pairs regardless of learning technique.
\modelname{}-3B achieves higher accuracy on all tasks, notably +12\% and +16\% on RoboSpatial, than tool-free SFT and RL respectively.

\Cref{fig:traces} shows qualitative examples.
We find that \modelname{} dynamically adapts its reasoning and tool-use strategies to each task. For example, it primarily relies on pointing for tasks such as spatial compatibility and relationship; it invokes depth estimation for relative-depth queries; and it composes multiple tools for more advanced reasoning like pose or grasp prediction. Moreover, \modelname{} has learned corrective behaviors, such as falling back to self-estimation when a tool fails, or switching to alternative pointing tools to refine uncertain detections.
Therefore, the model has \emph{learned internal procedures} for tool selection, ordering, and error recovery, rather than relying on hand-crafted pipelines in prior works \cite{lee2025apc,SpatialPIN_2024_arXiv}. 

\input{Tables/inference_robot_table}

\subsection{Experiments on Real Robot Manipulation}

In order to validate \modelname{} we conduct an experiment where robotics controls are presented as tools, see Figure~\ref{fig:robot_exp}.
The robot arm serves as an action tool, complementing vision-based perception tools.
By alternating between perception (pointing, segmentation, depth, grasp estimation) and action (capture, grasp) tools, 
the VLM orchestrates a closed perception–action loop fully guided by language reasoning, in contrast to prior work where robot action is an external process to model reasoning~\cite{song2025robospatial}.
We evaluate \modelname{}, Claude Sonnet~4.5, and GPT-5 in this tool-augmented system as well as comparing with a strong vision-language-action model, $\pi_{0.5}$~\cite{intelligence2025pi05visionlanguageactionmodelopenworld}. 
We focus on three type of tasks; pick, relational pick, and pick \& place, results from this experiments are presented in Table~\ref{tab:toolshed_realworld}. 
During the experiments we observed that \modelname{} is better grounded in spatial reasoning as well as being capable of orchestrating multiple tools, whereas 
other methods, like GPT-5, fail to chain tools coherently, sometimes inventing grasp poses or camera intrinsics instead of reusing computed values. 
Please consult 
the Appendix for further details.

\subsection{Ablation Study}

\input{Tables/ablation_table}

To analyze the contribution of each component in the \methodname{} framework, we perform systematic ablations on spatial reasoning benchmarks by removing (1) the \textit{IRL-trained teacher} (IRL-T), (2) the \textit{universal teacher} (Univ-T), and (3) the \textit{Stage~2 IRL} phase (S2-IRL).

In addition, to evaluate the importance of interactive reinforcement learning, we compare \methodname{} with two classic non-interactive training schemes: (a) \textit{Tool SFT} with the universal teacher, where the model is trained on multi-turn tool-use traces through direct supervision, and (b) \textit{Tool Non-Interactive Reinforcement Learning} (\textit{Tool NIRL}), which follows the conventional tool-learning setup in large language models (LLM) \cite{zhang2025nemotronresearchtooln1exploringtoolusinglanguage}. 
In Tool NIRL, ground-truth tool call traces are required, and the reward is based on the correctness of tool names, tool arguments, and answers.
Detailed configurations are provided in 
the Appendix. 

Quantitative results are in Table~\ref{tab:ablation_training}, with our main findings summarized:
(1) Removing the IRL-trained teacher leads to a sharp performance drop, particularly on tasks requiring fine spatial grounding such as RefSpatial and RoboSpatial. 
(2) Removing the universal teacher also degrades performance, especially on pose tasks that require multi-tool composition (\textit{e.g.}, segmentation + depth + 3D~bbox). 
(3) Stage~2 IRL provides the final boost of tool-augmented reasoning. Eliminating the Stage~2 IRL phase affects performance across RoboSpatial, RefSpatial, and pose tasks.
(4) Both Tool SFT and Tool NIRL baselines underperform \methodname{} by a large margin (+13.4 and +14.4 mean improvement, respectively). 
This suggests IRL is key to teaching VLMs reasoning over complex tool sequences.

%% file: Tables/inference_robot_table.tex

\begin{table}[h!]
\centering
\scriptsize
\caption{Real-world robotic manipulation performance of \modelname{} and zero-shot VLM baselines equipped with \systemname{}. Values are success rates (\%) for \emph{Pick} and \emph{Relation Pick} tasks, partial success rates (\%) for \emph{Pick \& Place}, and seconds for Time-to-First-Movement (TTFM).}

\begin{tabular}{@{}lcccc@{}} 
\toprule

& \multicolumn{3}{c}{\textbf{Real Robot Manipulation Tasks}} & \\ 
\cmidrule(lr){2-4}
\textbf{Model} & \textbf{Pick} & \textbf{Rel. Pick} & \textbf{Pick \& Place} & \textbf{TTFM} \\ 
\midrule
$\pi_{0.5}$ & \phantom{0}0 (0/7) & \phantom{0}0 (0/6) & \phantom{0}0 \phantom{0}(0/14) & \phantom{0}1s \\

GPT-5 + \systemname{} & 71 (5/7) & 33 (2/6) & 65 \phantom{0}(9/14) & 36s \\

Claude Sonnet 4.5 + \systemname{} & \textbf{86} (6/7) & 50 (3/6) & 79 (11/14) & 30s \\

Qwen2.5-VL-3B + \systemname{} & \phantom{0}0 (0/7) & \phantom{0}0 (0/6) & \phantom{0}0 \phantom{0}(0/14) & -\\

\textbf{\modelname{} (Ours)} & \textbf{86} (6/7) & \textbf{83} (5/6) & \textbf{86} (12/14) & 10s \\

\bottomrule
\end{tabular}
\label{tab:toolshed_realworld}
\end{table}

%% file: Tables/ablation_table.tex
\begin{table}[h!]
\caption{Ablation on training recipes.
IRL-T denotes the IRL-trained teacher; Univ-T denotes the universal (frontier-model) teacher; S2-IRL denotes the Stage-2 interactive RL phase.
Checkmarks indicate which components are included.}
\centering
\scriptsize

\begin{tabular}{@{}lccc cccc@{}}
\toprule
\textbf{Variant} & 
\textbf{IRL-T} & 
\textbf{Univ-T} &
\textbf{S2-IRL} &
\textbf{RoboSpatial} & 
\textbf{RefSpatial} & 
\textbf{Pose} & 
\textbf{Mean} \\
\midrule

\rowcolor[HTML]{F5F5F5}
\multicolumn{8}{c}{\textit{with Interactive RL}} \\

SpaceTools (Ours) & 
\gmark & \gmark & \gmark &
70.00 & 53.07 & 34.37 & 52.48 \\

w/o IRL Teacher & 
\rmark & \gmark & \gmark &
61.14 & 29.60 & 34.29 & 41.68 \\

w/o Univ. Teacher & 
\gmark & \rmark & \gmark &
65.14 & 54.51 & 8.92  & 42.86 \\

w/o Stage 2 IRL & 
\gmark & \gmark & \rmark &
67.71 & 51.98 & 33.28 & 50.99 \\
\midrule

\rowcolor[HTML]{F5F5F5}
\multicolumn{8}{c}{\textit{without Interactive RL}} \\

Tool SFT & 
\rmark & \gmark & \rmark &
59.71 & 24.91 & 32.94 & 39.19 \\

Tool NIRL & 
\rmark & \gmark & \rmark &
55.14 & 28.16 & 30.89 & 38.06 \\

\bottomrule
\end{tabular}
\label{tab:ablation_training}
\end{table}

%% file: tex/6_discussion.tex
\vspace{-.1in}
\section{Discussion \& Conclusion}
\vspace{-.1in}
\input{Tables/inference_table}

Agentic VLMs hold the promise of reasoning through arbitrary external tools.
Motivated by this, we examine whether large VLMs can improve their spatial reasoning by leveraging vision tools in a fully zero-shot setting.
As shown in \Cref{tab:toolshed_comparison}, tool integration yields clear gains on tasks requiring precise spatial grounding or explicit geometric reasoning. 
For example, GPT-5 with \systemname{} improves on RefSpatial (from 23.1 to 36.1) and pose estimation (from 9.0 to 15.0), 
suggesting that tool feedback mitigates limitations in spatial grounding and 3D understanding. 
In contrast, high-level tasks such as RoboSpatial and BLINK show mixed trends, as models tend to overuse tools and struggle to correctly interpret nuanced tool outputs.
We also find that IRL improves out-of-domain generalization. When a model is trained to use a single powerful tool such as pointing~\cite{zhou2025roborefer}, it not only performs better on its in-domain benchmark but also transfers unexpectedly well.
For instance, a model trained only on RoboSpatial~\cite{song2025robospatial} reaches 72.3\% accuracy on that benchmark and still achieves 34.3\% on RefSpatial—where other fine-tuning approaches score zero.
These results highlight the promise of agentic VLMs and their ability to acquire new skills through tool use.

In conclusion, we introduce \methodname{}, a new method for training tool-augmented VLMs through progressive and interactive learning. To support this, we built Toolshed, a system for deploying diverse tools at scale for online interaction during training. Our experiments show that our trained model, \modelname{}, achieves state-of-the-art performance on multiple spatial reasoning benchmarks and exhibits strong out-of-distribution generalization, including the ability to use a robot as a tool. This work demonstrates that VLMs can acquire complex spatial reasoning capabilities through learned tool coordination rather than architectural modification or large-scale data-driven fine-tuning.

%% file: Tables/inference_table.tex





\begin{table}[h!]
\centering
\scriptsize
\caption{Comparison of proprietary models with and without the \systemname{} enhancement across robotic spatial reasoning benchmarks. Values are normalized accuracy (\%).
}
\begin{tabular}{@{}lccccc@{}} 
\toprule
\multirow{2}{*}{\textbf{Model}} 
& \multirow{2}{*}{\textbf{RoboSpatial}} 
& \multirow{2}{*}{\textbf{BLINK}} 
& \multirow{2}{*}{\textbf{RefSpatial}} 
& \multicolumn{2}{c}{\textbf{BOP-ASK}} \\ 
\cmidrule(lr){5-6}
 &  &  &  & \textbf{Pose} & \textbf{Grasp (MACE)} \\ 
\midrule
GPT-5 & 58.39 & 66.13 \phantom{0} & 23.10 \phantom{0} & \phantom{0}9.03 \phantom{0} & 39.59 \phantom{0}  \\
\hspace{1em}+ \systemname{} & 55.14 & 90.32 $\uparrow$ & 36.10 $\uparrow$ & 15.00 $\uparrow$ & 41.49 $\uparrow$ \\
\cmidrule(lr){1-1} 

Claude & 57.43 & 78.23 \phantom{0}  & \phantom{0}7.49\phantom{0} & 1.67 \phantom{0} & 40.12 \phantom{0} \\
\hspace{1em}+ \systemname{} & 52.86 & 75.00 \phantom{0} & 27.80 $\uparrow$ & 25.00 $\uparrow$ & 44.19 $\uparrow$ \\
\bottomrule
\end{tabular}
\label{tab:toolshed_comparison}
\end{table}

%% file: tex/appendix/0_main.tex
\section*{Supplementary Material}

\noindent We provide additional details and extended results in the supplementary materials:

\begin{itemize}
    \item Appendix~\ref{app-sec:limitations}: Limitations and future directions.
    \item Appendix~\ref{sec:toolshed_details}: Further details on the Toolshed system.
    \item Appendix~\ref{app-sec:method}: Expanded method descriptions.
    \item Appendix~\ref{sec:app-implementation-details}: Additional implementation details.
    \item Appendix~\ref{sec:app-exp}: Extended experimental results.
\end{itemize}

\input{tex/appendix/1_limitations}

\input{tex/appendix/2_toolshed}
\input{tex/appendix/3_method}
\input{tex/appendix/4_implementation}
\input{tex/appendix/5_results}

%% file: tex/appendix/1_limitations.tex
\section{Limitations and Future Directions}
\label{app-sec:limitations}

Our work shows that tool-augmented spatial reasoning, enabled through \methodname{} and the Toolshed infrastructure, provides an effective and scalable foundation for training VLMs with strong spatial reasoning, robust tool coordination, and broad generalization across diverse tasks and embodiments.
At the same time, this framework opens several promising directions that fall beyond our present scope but merit deeper exploration. We discuss these limitations and future opportunities below.

\paragraph{Application scope.} A natural next step is to broaden the range of tasks and environments in which tool-augmented spatial reasoning is applied. Our current scope focuses on short- or medium-horizon tasks, such as spatial question answering or grasp-and-place manipulations. Extending to more complex, longer-horizon, or multi-stage tasks may further reveal the potential of tool-augmented reasoning, allowing the model to concentrate on reasoning and decision-making rather than learning numerous precise perceptual subtasks. Moreover, integrating richer environments, including large-scale robotic simulation, interactive game environments, or physics-rich virtual worlds, could support more diverse experiences and ultimately more general embodied spatial intelligence.

\paragraph{Methodology.} From a methodological perspective, several directions could strengthen the flexibility and robustness of tool-augmented spatial reasoning. Although Toolshed supports image-level tool outputs, this work primarily explores tools that return structured text or variables (\textit{e.g.}, point cloud). Extending the model to reason over visual outputs from tools may unlock more expressive or fine-grained reasoning behaviors. Another important direction is systematically improving how VLMs perceive, verify, and recover from tool errors or inaccuracies. Moreover, under a modular perspective, future work could investigate enhancing particular system capabilities by upgrading a single tool without modifying other components, while ensuring that overall tool coordination remains robust. Additionally, while our RL exploration focuses on prompt design, loss functions, and task-specific reward formulations, alternative RL approaches, such as stepwise reward formulations, may improve learning effectiveness in large multi-tool action spaces. Finally, continual learning of new tools is also an important future direction.

\paragraph{Infrastructure.} On the system side, Toolshed provides a scalable backbone for interactive tool use and learning, but there remains room to further improve its efficiency and resource utilization. Serving many heterogeneous tools in parallel can introduce latency and memory bottlenecks, particularly for high-resolution vision tools or robot-in-the-loop executions. In this work, we mitigate the latter by using mock robot tools during training. Future advances in scheduling, caching, batching, and asynchronous execution could potentially enhance performance and even support real robot execution effectively during interactive learning. Additionally, developing lighter-weight tools, model-side approximators, or memory-optimized deployment strategies may reduce overhead and enable larger-scale training or more complex task environments. These improvements would allow the framework to scale more gracefully as tool diversity and task complexity increase.

\begin{figure}[h]
    \centering
    \includegraphics[width=0.6\linewidth]{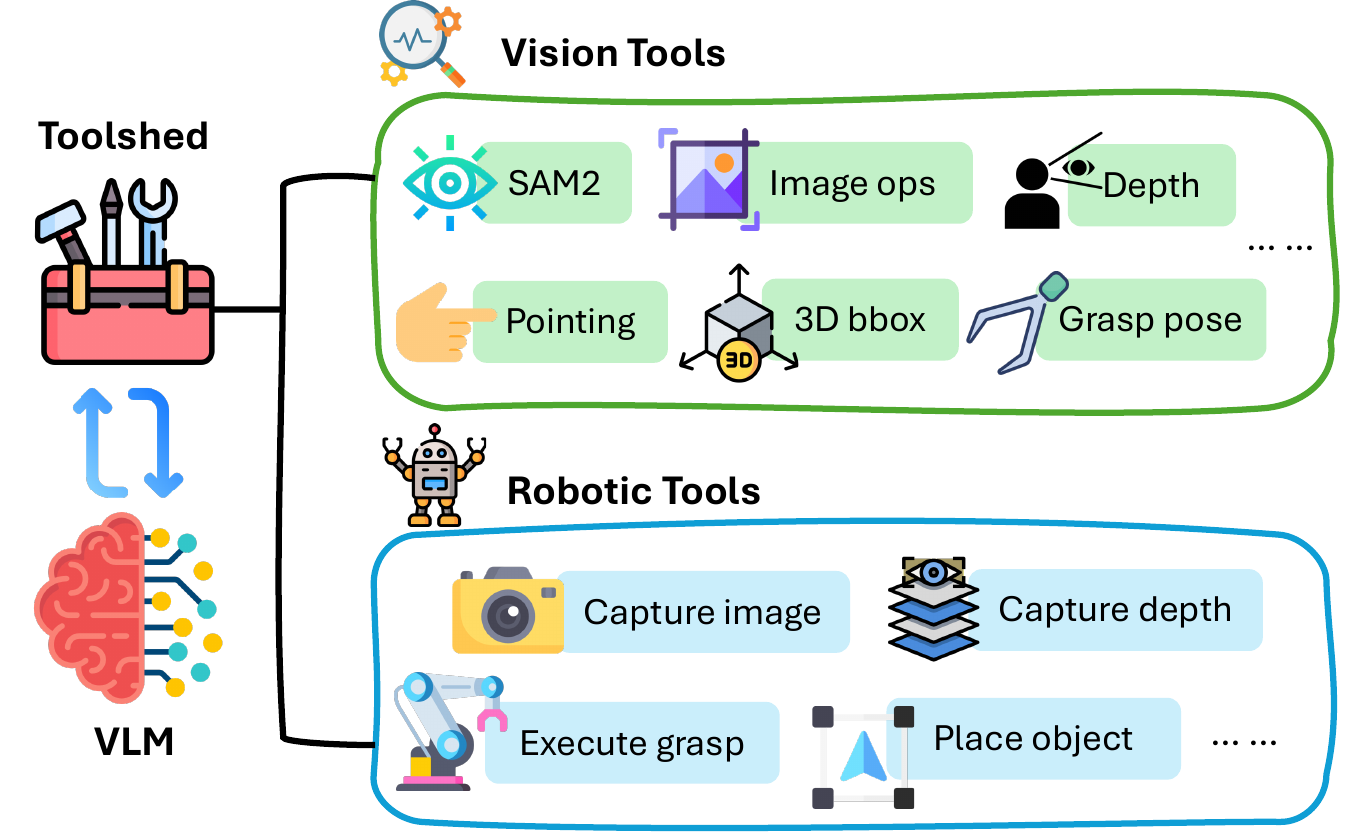}
    \caption{The \textbf{\systemname{}} infrastructure linking a VLM with modular vision and robotic tools under a unified toolbox for perception and control.
    }
    \label{fig:toolbox}
\end{figure}

%% file: tex/appendix/2_toolshed.tex
\subsection{System Design}
This work aims to enable learning and inference with multiple interactive vision tools for spatial reasoning. 
Effective tool-augmented spatial reasoning requires multi-turn, state-dependent communication between the VLM and its vision tools. However, many essential tools such as object detectors, depth estimators, and 3D reconstruction modules, are computationally heavy and often dominate both inference and training time in VLM–RL pipelines. Moreover, modern training relies on batched generation, where multiple conversations are executed in parallel. In naïve implementations, a single blocking tool call can stall the entire batch, effectively reducing tool interaction to a serial process. This makes it crucial to keep tools continuously available, pre-loaded on device, and capable of serving multiple concurrent conversations. To address these challenges, we introduce \textbf{\systemname{}} (visualized in Figure~\ref{fig:toolbox}), a distributed toolkit that enables scalable, asynchronous, and parallel vision tool interaction.

\begin{itemize}
    \item \textbf{Decoupled execution.} Tool invocations run independently from the policy's main inference loop, avoiding blocking calls that would otherwise stall unrelated computations.
    \item \textbf{Asynchronous Processing.} Multiple parallel instances can serve the same tool, each receiving inputs and producing outputs asynchronously, enabling high throughput even under large-scale rollouts.
    \item \textbf{Resources isolation.} Tool instances are assigned isolated resources based on the computational profile of each tool.
    \item \textbf{Environment isolation.} Each tool type is hosted in a dedicated python environment, solving the dependency compatibility issue that comes with hosting multiple computer vision tools in a single system.
    \item \textbf{Elastic scaling.} The system design supports automatic spawning of additional tool workers in response to bursts of tool usage, allowing throughput to remain stable even for large batch rollouts. (This capability is part of the infrastructure design but was not enabled in our training experiments.)
    \item \textbf{Multimodal data passing.} Seamless exchange of text, images, and structured variables (\textit{e.g.}, 3D point clouds) is supported between the VLM and tools, even when they run on different devices or GPU nodes. This enables tool workflows that require different types of inputs and outputs.
\end{itemize}

In practice, \systemname{} is implemented on top of the Ray\textsuperscript{3}\label{fn:toolshed} distributed execution framework, which provides lightweight task scheduling, actor management, and high-throughput message passing. For interactive reinforcement learning, \systemname{} integrates seamlessly with VERL\textsuperscript{4}: VERL's asynchronous multi-turn rollouts align naturally with \systemname{}'s asynchronous tool actors, enabling us to parallelize expensive perception, generation, and simulation steps without slowing down rollouts. This results in significantly higher steps-per-second compared to monolithic or synchronous training setups. For inference, \systemname{} can be attached both to our trained model and to proprietary models (e.g., GPT-5, Claude) via simple API calls, enhancing their robotic spatial reasoning capabilities.

\subsection{Provided Tools}
\paragraph{Vision tools.}
We provide the following vision tools. \texttt{image\_ops} offers basic image manipulations such as point- and mask-based indexing. \texttt{sam2} performs instance segmentation from one or more clicks, powered by Segment Anything~2~\cite{ravi2024sam2}. \texttt{point1} and \texttt{point2} are two object-pointing detectors backed by RoboRefer~\cite{zhou2025roborefer} and Molmo~\cite{Deitke2025Molmo}, respectively. \texttt{depth\_estimator} predicts monocular depth and reconstructs 3D point clouds using DepthPro~\cite{depthpro2025}. \texttt{compute\_bbox} estimates 3D bounding boxes and object poses from reconstructed geometry, while \texttt{compute\_grasp} predicts collision-free grasp poses for robotic manipulation. Finally, \texttt{code\_executor} allows the VLM to execute small Python snippets for orchestrating multi-tool workflows, returning results with captured \texttt{stdout}/\texttt{stderr} and optionally caching intermediate outputs for reuse.

\paragraph{Robotic tools.}
We integrate a set of robotic tools that enable embodied perception and manipulation.
\texttt{capture\_image} captures RGB observations from the robot's onboard camera and stores them for subsequent visual processing.
\texttt{capture\_depth} acquire depth information from the scene, returning a depth map with focal length and a full 3D point cloud reconstruction.
\texttt{execute\_grasp} executes a grasp given a 4$\times$4 transformation matrix representing the target grasp pose and reports execution success and timing feedback.
\texttt{place\_object} places an object at a specified 2D image coordinate, confirming successful placement in the returned text message. Apart from the above real robotic tools that control real-world robot arms, we also provide a set of \texttt{mock\_robot} tools without relying on real robots for the ease of data generation and training.
Together, these tools provide a physical interface that allows the VLM to perceive, reason, and act within the real world, enabling unified spatial reasoning and robotic control.

\subsection{Example Process of Launching Tool IRL}

We show a condensed workflow of launching tools and interactive reinforcement learning 
via a single bash script in Listing~\ref{lst:toolshed_workflow}.

\begin{lstlisting}[language=bash,
caption={Example workflow for launching Toolshed and running VERL GRPO training with tool calling enabled.},
label={lst:toolshed_workflow}]
# -------------------------------------
# Graceful cleanup (kills the Toolshed actor process so detached actors vanish)
# -------------------------------------
TOOLSHED_PID=""
cleanup() {
  echo "Cleaning up"
  if [[ -n "$TOOLSHED_PID" ]]; then
    echo "Killing(PID=$TOOLSHED_PID)"
    kill $TOOLSHED_PID 2>/dev/null
    wait $TOOLSHED_PID 2>/dev/null
    echo "Stopped."
  fi
  exit 0
}
trap cleanup SIGINT SIGTERM EXIT

# 1. Launch Toolshed with GPU-backed vision tools
python - <<'PY'
import ray
from toolshed import start_toolkit

ray.init(address='auto')

tool_configs = {
    "point1": {"num_actors": 2, "resources": {"num_gpus": 0.5}},
    "point2": {"num_actors": 2, "resources": {"num_gpus": 0.5}},
    "sam2": {"num_actors": 4, "resources": {"num_gpus": 0.2}},
    "depth_estimator": {"num_actors": 4, "resources": {"num_gpus": 0.2}}
    ...
}

pg = ray.util.placement_group([{"CPU": 8, "GPU": 8}], strategy="STRICT_PACK")
ray.get(pg.ready())
router = start_toolkit(tool_configs, dashboard=True, placement_group=pg)
print("Started:", list(tool_configs))
PY

# 2. Generate a YAML tool config and make sure tools are ready
python generate_toolshed_config.py --output toolshed_config.yaml

# 3. Launch VERL GRPO training with Toolshed Integration
python -m verl.trainer.main_ppo \
    actor_rollout_ref.model.path=Qwen/Qwen2.5-VL-3B-Instruct \
    ...
    actor_rollout_ref.rollout.multi_turn.tool_config_path=toolshed_config.yaml

echo "Finished." 
\end{lstlisting}

\section{Details of \systemname{}}
\label{sec:toolshed_details}

\subsection{Example Toolshed Integration with Proprietary Models}

\begin{figure}[h]
\centering
\includegraphics[width=\linewidth]{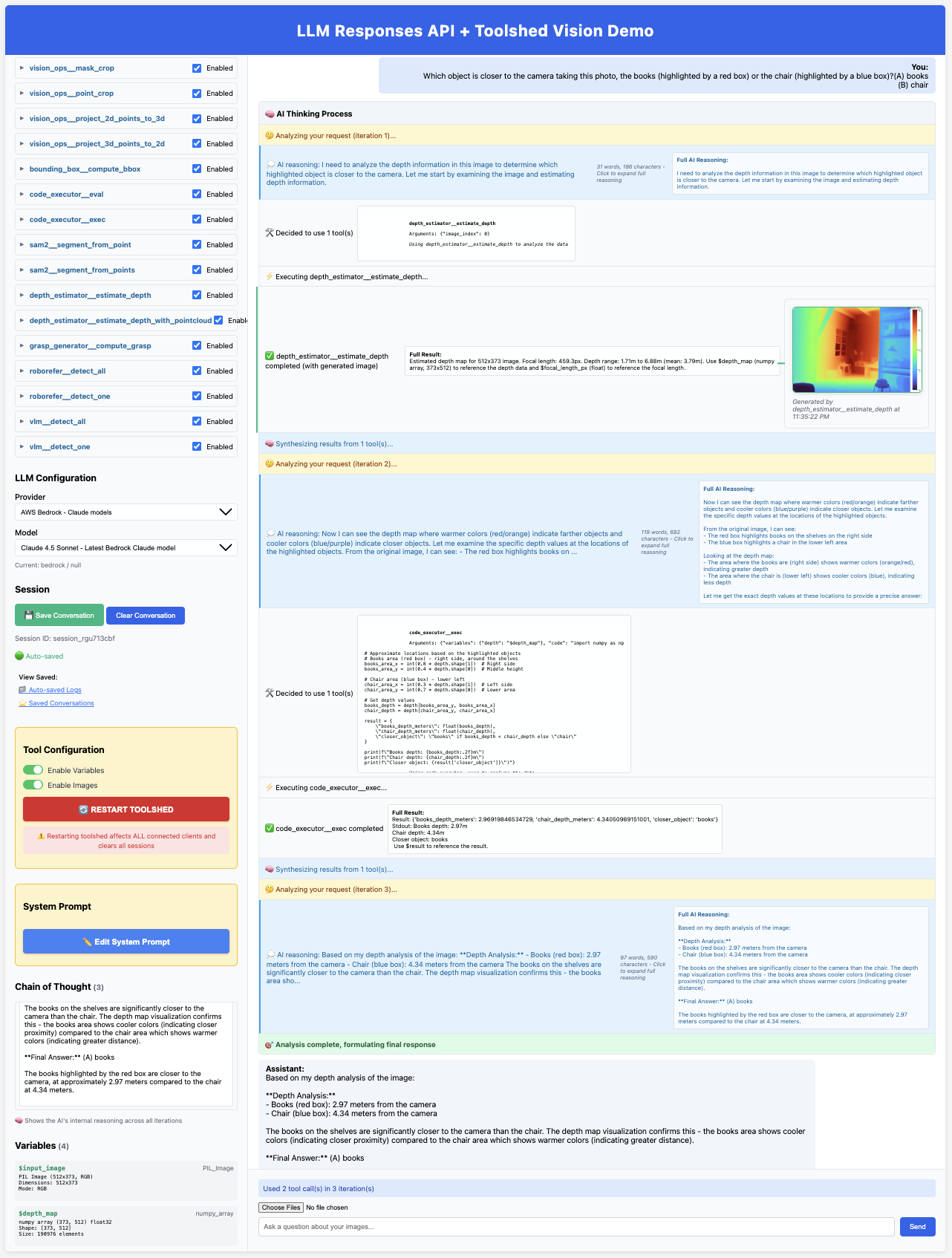}
\caption{Interactive web demo illustrating Claude's tool-augmented reasoning when integrated with Toolshed.}
\label{fig:claude_demo}
\end{figure}

Figure~\ref{fig:claude_demo} illustrates our integration of Toolshed with a proprietary model (Claude Sonnet~4.5), enabling seamless interactive communication among the user, external tools, and the VLM.

\clearpage

\subsection{Detailed Tool APIs}
\label{app-sec:toolAPIs}

Toolshed uses pythonic, modular interfaces \& APIs (Appx B), \textit{minimizing} new tool integration effort: only definition with \texttt{@tool\_method} decorator is required; execution, RL integration, etc, are handled \textit{automatically}. Complete API of the computer vision and robotic tools supported by Toolshed include:

\FloatBarrier
\begin{description}[style=nextline]
\item[\texttt{image\_ops.point\_crop(data, x, y)}]
  \textbf{Purpose:} Get the data value in the numpy ndarray `data` at the given coordinate.
    \begin{itemize}
    \item \textbf{Inputs}:
      \begin{itemize}
        \item \texttt{data}: Numpy ndarray of shape (H, W) or (H, W, C), or PIL Image
        \item \texttt{x}: Normalized x-coordinate, float in \([0,1]\)
        \item \texttt{y}: Normalized y-coordinate, float in \([0,1]\)
      \end{itemize}
    \item \textbf{Outputs}:
      \begin{itemize}
        \item \textbf{Raw value}: \texttt{data} value indexed at the input coordinate.
        \item \textbf{Text}: Reports the information about the pixel value at the given coordinates.
      \end{itemize}
    \end{itemize}

  \item[\texttt{image\_ops.point\_crop(image, points)}]
  Crop image to minimally encompass all given points.
  \begin{itemize}
    \item \textbf{Inputs}:
      \begin{itemize}
        \item \texttt{image}: PIL.Image
        \item \texttt{points}: list of \((x,y)\) normalized floats in \([0,1]\)
      \end{itemize}
    \item \textbf{Outputs}:
      \begin{itemize}
        \item \textbf{Raw value}: \texttt{PIL.Image} cropped to the bounding box of the points
        \item \textbf{Text}: Reports the crop box, size, and number of points
        \item \textbf{Image}: The cropped region only (no overlay)
        \item \textbf{Variables}: \texttt{cropped\_image} (\texttt{PIL.Image})
      \end{itemize}
  \end{itemize}

  \item[\texttt{image\_ops.mask\_crop(image, mask)}]
  Crop to the tight bounding box of a boolean mask; outside-mask pixels are set to white.
  \begin{itemize}
    \item \textbf{Inputs}:
      \begin{itemize}
        \item \texttt{image}: PIL.Image
        \item \texttt{mask}: boolean \texttt{numpy.ndarray} of shape \(H \times W\) matching the image
      \end{itemize}
    \item \textbf{Outputs}:
      \begin{itemize}
        \item \textbf{Raw value}: \texttt{PIL.Image} of the masked region on white background, cropped to mask bounds
        \item \textbf{Text}: Reports crop box, size, and mask coverage percentage
        \item \textbf{Image}: Masked crop on white background (no overlay)
        \item \textbf{Variables}: \texttt{masked\_crop} (\texttt{PIL.Image})
      \end{itemize}
  \end{itemize}

  \item[\texttt{sam2.segment\_from\_point(image, x, y)}]
  Segment the object at a single pixel coordinate.
  \begin{itemize}
    \item \textbf{Inputs}:
      \begin{itemize}
        \item \texttt{image}: PIL.Image
        \item \texttt{x}: Normalized x-coordinate, float in \([0,1]\)
        \item \texttt{y}: Normalized y-coordinate, float in \([0,1]\)
      \end{itemize}
    \item \textbf{Outputs}:
      \begin{itemize}
        \item \textbf{Raw value}: \texttt{dict} with \texttt{mask} (boolean \(H{\times}W\) \texttt{numpy.ndarray}) and \texttt{iou\_score} (float)
        \item \textbf{Text}: Reports the click location and IoU score
        \item \textbf{Image}: Original image with semi-transparent green mask, a white mask outline, and a red circular point marker (white outline)
        \item \textbf{Variables}: \texttt{segmentation\_mask} (boolean \(H{\times}W\) \texttt{numpy.ndarray})
      \end{itemize}
  \end{itemize}

  \item[\texttt{sam2.segment\_from\_points(image, points)}]
  Segment an object using multiple foreground points.
  \begin{itemize}
    \item \textbf{Inputs}:
      \begin{itemize}
        \item \texttt{image}: PIL.Image
        \item \texttt{points}: list of \((x,y)\) normalized floats in \([0,1]\)
      \end{itemize}
    \item \textbf{Outputs}:
      \begin{itemize}
        \item \textbf{Raw value}: \texttt{dict} with \texttt{mask} (boolean \(H{\times}W\) \texttt{numpy.ndarray}) and \texttt{iou\_scores} (1-D \texttt{numpy.ndarray})
        \item \textbf{Text}: Reports the number of points and the best IoU score
        \item \textbf{Image}: Original image with semi-transparent green mask, a white mask outline, and red circular markers at all provided points.
        \item \textbf{Variables}: \texttt{segmentation\_mask} (boolean \(H{\times}W\) \texttt{numpy.ndarray})
      \end{itemize}
  \end{itemize}

  \item[\texttt{point1.detect\_one(image, obj\_name)}]
  Identify one instance of the named object by pointing to them with Roborefer.
  \begin{itemize}
    \item \textbf{Inputs}:
      \begin{itemize}
        \item \texttt{image}: PIL.Image
        \item \texttt{obj\_name}: string
      \end{itemize}
    \item \textbf{Outputs}:
      \begin{itemize}
        \item \textbf{Raw value}: String serialization of a  normalized point coordinate \((x,y) \in [0,1]^2\) 
        \item \textbf{Text}: Reports the object name, count, and the normalized point
        \item \textbf{Image}: Original image with red circular point markers at detected locations (white outlines)
        \item \textbf{Variables}: \texttt{\textit{<obj\_name>}\_detection} (\((x,y)\) floats in \([0,1]\); spaces in \textit{obj\_name} replaced with underscores)
      \end{itemize}
  \end{itemize}

  \item[\texttt{point1.detect\_all(image, obj\_name)}]
  Identify instances of the named object by pointing to them with Roborefer.
  \begin{itemize}
    \item \textbf{Inputs}:
      \begin{itemize}
        \item \texttt{image}: PIL.Image
        \item \texttt{obj\_name}: string
      \end{itemize}
    \item \textbf{Outputs}:
      \begin{itemize}
        \item \textbf{Raw value}: String serialization of a list of normalized point coordinates \((x,y) \in [0,1]^2\) 
        \item \textbf{Text}: Reports the object name, count, and the list of normalized points
        \item \textbf{Image}: Original image with red circular point markers at detected locations (white outlines)
        \item \textbf{Variables}: \texttt{\textit{<obj\_name>}\_detections} (list of \((x,y)\) floats in \([0,1]\); spaces in \textit{obj\_name} replaced with underscores)
      \end{itemize}
  \end{itemize}

  \item[\texttt{point2.detect\_one(image, obj\_name)}]
  Identify one instance of the named object by pointing to them with Molmo.
  \begin{itemize}
    \item \textbf{Inputs}:
      \begin{itemize}
        \item \texttt{image}: PIL.Image
        \item \texttt{obj\_name}: string
      \end{itemize}
    \item \textbf{Outputs}:
      \begin{itemize}
        \item \textbf{Raw value}: String serialization of a  normalized point coordinate \((x,y) \in [0,1]^2\) 
        \item \textbf{Text}: Reports the object name, count, and the normalized point
        \item \textbf{Image}: Original image with red circular point markers at detected locations (white outlines)
        \item \textbf{Variables}: \texttt{\textit{<obj\_name>}\_detection} (\((x,y)\) floats in \([0,1]\); spaces in \textit{obj\_name} replaced with underscores)
      \end{itemize}
  \end{itemize}

  \item[\texttt{point2.detect\_all(image, obj\_name)}]
  Identify instances of the named object by pointing to them with Molmo.
  \begin{itemize}
    \item \textbf{Inputs}:
      \begin{itemize}
        \item \texttt{image}: PIL.Image
        \item \texttt{obj\_name}: string
      \end{itemize}
    \item \textbf{Outputs}:
      \begin{itemize}
        \item \textbf{Raw value}: String serialization of a list of normalized point coordinates \((x,y) \in [0,1]^2\) 
        \item \textbf{Text}: Reports the object name, count, and the list of normalized points
        \item \textbf{Image}: Original image with red circular point markers at detected locations (white outlines)
        \item \textbf{Variables}: \texttt{\textit{<obj\_name>}\_detections} (list of \((x,y)\) floats in \([0,1]\); spaces in \textit{obj\_name} replaced with underscores)
      \end{itemize}
  \end{itemize}

  \item[\texttt{depth\_estimator.estimate\_depth(image)}]
  Monocular depth estimation with DepthPro.
  \begin{itemize}
    \item \textbf{Inputs}:
      \begin{itemize}
        \item \texttt{image}: PIL.Image
      \end{itemize}
    \item \textbf{Outputs}:
      \begin{itemize}
        \item \textbf{Raw value}: \texttt{dict} with \texttt{depth\_map} (\(H{\times}W\) float array, meters), \texttt{focal\_length\_px} (float), \texttt{width} (int), \texttt{height} (int)
        \item \textbf{Text}: Reports image size, focal length, and depth range statistics
        \item \textbf{Image}: Colorized depth map with a vertical scale bar on the right labeled "Depth (m)"
        \item \textbf{Variables}: \texttt{depth\_map} (\(H{\times}W\) float array), \texttt{focal\_length\_px} (float)
      \end{itemize}
  \end{itemize}

  \item[\texttt{depth\_estimator.\\estimate\_depth\_with\_pointcloud(image)}]
  Monocular depth estimation and 3D point cloud generation with DepthPro.
  \begin{itemize}
    \item \textbf{Inputs}:
      \begin{itemize}
        \item \texttt{image}: PIL.Image
      \end{itemize}
    \item \textbf{Outputs}:
      \begin{itemize}
        \item \textbf{Raw value}: \texttt{dict} with \texttt{depth\_map} (\(H{\times}W\) float array, meters),  \texttt{point\_cloud} (\(N{\times}3\) float array of 3D points in camera coordinates), \texttt{focal\_length\_px} (float), \texttt{width} (int), \texttt{height} (int)
        \item \textbf{Text}: Reports image size, focal length, and depth range statistics
        \item \textbf{Image}: Colorized depth map with a vertical scale bar on the right labeled "Depth (m)"
        \item \textbf{Variables}: \texttt{depth\_map} (\(H{\times}W\) float array), \texttt{point\_cloud} (\(N{\times}3\) float array of 3D points in camera coordinates), \texttt{focal\_length\_px} (float)
      \end{itemize}
  \end{itemize}

  \item[\texttt{grasp\_generator.compute\_grasp(point\_cloud, mask, image, focal\_length\_px)}]
  Generate a single grasp pose for a masked subset of a point cloud with GraspGen.
  \begin{itemize}
    \item \textbf{Inputs}:
      \begin{itemize}
        \item \texttt{point\_cloud}: \(N{\times}3\) numpy float array
        \item \texttt{mask}: Boolean segmentation mask
        \item \texttt{image}: PIL.Image
        \item \texttt{focal\_length\_px}: float
      \end{itemize}
    \item \textbf{Outputs}:
      \begin{itemize}
        \item \textbf{Raw value}: Collision-free grasp pose, and  collision-free confidence
        \item \textbf{Text}: Reports the collision-free grasp confidence, the total number of generated grasps and the percentage of collision-free grasps, and the projected 2D gripper points of the best grasp pose in normalized coordinates
        \item \textbf{Image}: Original image overlayed with projected X-(red), Y-(green), Z-(blue) gripper axes
        \item \textbf{Variables}: \texttt{grasp\_pose} ($4\times4$ ndarray, OpenCV camera frame)
      \end{itemize}
  \end{itemize}

  \item[\texttt{3d\_bbox.compute\_bbox(point\_cloud, mask, focal\_length\_px)}]
  Compute an oriented bounding box for a masked subset of a point cloud.
  \begin{itemize}
    \item \textbf{Inputs}:
      \begin{itemize}
        \item \texttt{point\_cloud}: \(N{\times}3\) numpy float array
        \item \texttt{mask}: Boolean segmentation mask
        \item \texttt{focal\_length\_px}: float
      \end{itemize}
    \item \textbf{Outputs}:
      \begin{itemize}
        \item \textbf{Raw value}: Box corners in 3D, box corners in 2D, edges, and extent
        \item \textbf{Text}: Summary containing number of input points, the point coordinates in 3d and 2d,
        mask shape, box extents, and edges
        \item \textbf{Image}: No image output
        \item \textbf{Variables}: \texttt{obb\_corners\_3d} ($8\times3$ list of lists, meters in opencv camera frame), \texttt{obb\_corners\_2d} ($8\times2$ list of lists, normalized image coordinates), \texttt{extent} (3-element ndarray, extent of the bounding box in meters), \texttt{edges} (list of pairs of integers, edges of the bounding box defined by the indices of the corners)
      \end{itemize}
  \end{itemize}

\item[\texttt{code\_executor.exec(code)}]
  Execute a multi-line Python block (imports limited to \texttt{math} and \texttt{numpy}).
  \begin{itemize}
    \item \textbf{Inputs}:
      \begin{itemize}
        \item \texttt{code}: string
      \end{itemize}
    \item \textbf{Outputs}:
      \begin{itemize}
        \item \textbf{Raw value}: tuple \((\textit{result}, \textit{stdout}, \textit{stderr})\)
        \item \textbf{Text}: Summarizes the result, captured stdout, and stderr; notes a stored variable if applicable
        \item \textbf{Image}: No image output
        \item \textbf{Variables}: \texttt{last\_exec\_result} (present iff a non-\texttt{None} result and variables are enabled)
      \end{itemize}
  \end{itemize}

  \item[\texttt{code\_executor.eval(expression)}]
  Evaluate a single Python expression.
  \begin{itemize}
    \item \textbf{Inputs}:
      \begin{itemize}
        \item \texttt{expression}: string
      \end{itemize}
    \item \textbf{Outputs}:
      \begin{itemize}
        \item \textbf{Raw value}: tuple \((\textit{result}, \textit{stdout}, \textit{stderr})\)
        \item \textbf{Text}: Summarizes the result, captured stdout, and stderr; notes a stored variable if applicable
        \item \textbf{Image}: No image output
        \item \textbf{Variables}: \texttt{last\_eval\_result} (present iff a non-\texttt{None} result and variables are enabled)
      \end{itemize}
  \end{itemize}

  \item[\texttt{mock\_robot.capture\_image(mock\_data)}]
  Return the mock image from the dataset without a real robot.
  \begin{itemize}
    \item \textbf{Inputs}:
      \begin{itemize}
        \item \texttt{mock\_data}: \texttt{dict} with \texttt{mock\_image}
      \end{itemize}
    \item \textbf{Outputs}:
      \begin{itemize}
        \item \textbf{Raw value}: Image from mock camera
        \item \textbf{Text}: Image dimensions and capture status
        \item \textbf{Image}: Captured image from mock camera
        \item \textbf{Variables}: \texttt{captured\_image} (PIL Image)
      \end{itemize}
  \end{itemize}

  \item[\texttt{mock\_robot.get\_depth(mock\_data)}]
  Return the mock depth from the dataset without a real robot.
  \begin{itemize}
    \item \textbf{Inputs}:
      \begin{itemize}
        \item \texttt{mock\_data}: \texttt{dict} with \texttt{mock\_image} (PIL Image), \texttt{mock\_depth\_map} (numpy array), \texttt{mock\_focal\_length\_px} (float), \texttt{image\_width} (int), \texttt{image\_height} (int)
      \end{itemize}
    \item \textbf{Outputs}:
      \begin{itemize}
        \item \textbf{Raw value}: \texttt{image} (PIL Image), \texttt{depth\_map} (numpy array), \texttt{focal\_length\_px} (float), \texttt{width} (int), \texttt{height} (int)
        \item \textbf{Text}: Summary of depth data including image dimensions, focal length, and depth statistics
        \item \textbf{Image}: A colorized depth map visualization where closer objects appear cooler (blue/purple) and distant objects appear warmer (red/yellow)
        \item \textbf{Variables}: \texttt{depth\_map} (2D numpy array of depth values in meters), \texttt{foca\_length\_px} (float, estimated focal length in pixels)
      \end{itemize}
  \end{itemize}
  \item[\texttt{mock\_robot.\\get\_depth\_with\_pointcloud(mock\_data)}]
  Return the mock depth and and point cloud generation from the dataset without a real robot.
  \begin{itemize}
    \item \textbf{Inputs}:
      \begin{itemize}
        \item \texttt{mock\_data}: \texttt{dict} with \texttt{mock\_image} (PIL Image), \texttt{mock\_depth\_map} (numpy array), 
        \texttt{mock\_point\_cloud} (numpy array),
        \texttt{mock\_focal\_length\_px} (float), \texttt{image\_width} (int), \texttt{image\_height} (int).
      \end{itemize}
    \item \textbf{Outputs}:
      \begin{itemize}
        \item \textbf{Raw value}: \texttt{image} (PIL Image), \texttt{depth\_map} (numpy array), 
        \texttt{mock\_point\_cloud} (numpy array),
        \texttt{focal\_length\_px} (float), \texttt{width} (int), \texttt{height} (int).
        \item \textbf{Text}: Summary of depth data and and point cloud generation including image dimensions, focal length, depth statistics, and point cloud size
        \item \textbf{Image}: A colorized depth map visualization where closer objects appear cooler (blue/purple) and distant objects appear warmer (red/yellow)
        \item \textbf{Variables}: \texttt{depth\_map} (2D numpy array of depth values in meters), 
        \texttt{point\_cloud} ($N\times3$ array of 3D points),
        \texttt{foca\_length\_px} (float, estimated focal length in pixels)
      \end{itemize}
  \end{itemize}

  \item[\texttt{mock\_robot.execute\_grasp(grasp\_pose)}]
  Simulate executing a grasp (always succeeds).
  \begin{itemize}
    \item \textbf{Inputs}:
      \begin{itemize}
        \item \texttt{grasp\_pose}: $4\times4$ transformation matrix representing the grasp pose in the robot's camera frame, OpenCV convention
      \end{itemize}
    \item \textbf{Outputs}:
      \begin{itemize}
        \item \textbf{Raw value}: \texttt{success} (boolean), \texttt{execution\_time\_s} (float)
        \item \textbf{Text}: Confirmation that grasp was successful
        \item \textbf{Image}: No image output
        \item \textbf{Variables}: No variable output
      \end{itemize}
  \end{itemize}
  \item[\texttt{mock\_robot.\\place\_object\_at\_2d\_location(point\_2d)}]
  Simulate placing object at 2D location (always succeeds).
  \begin{itemize}
    \item \textbf{Inputs}:
      \begin{itemize}
        \item \texttt{point\_2d}: 2D normalized image coordinate where the object should be placed
      \end{itemize}
    \item \textbf{Outputs}:
      \begin{itemize}
        \item \textbf{Raw value}: \texttt{success} (boolean), \texttt{execution\_time\_s} (float)
        \item \textbf{Text}: Confirmation that placement was successful
        \item \textbf{Image}: No image output
        \item \textbf{Variables}: No variable output
      \end{itemize}
  \end{itemize}

  \item[\texttt{mock\_robot.\\place\_object\_at\_3d\_location(point\_3d)}]
  Simulate placing object at 3D location (always succeeds).
  \begin{itemize}
    \item \textbf{Inputs}:
      \begin{itemize}
        \item \texttt{point\_3d}: 3D point in the robot's camera frame (list or numpy array) where the object should be placed
      \end{itemize}
    \item \textbf{Outputs}:
      \begin{itemize}
        \item \textbf{Raw value}: \texttt{success} (boolean), \texttt{execution\_time\_s} (float)
        \item \textbf{Text}: Confirmation that placement was successful
        \item \textbf{Image}: No image output
        \item \textbf{Variables}: No variable output
      \end{itemize}
  \end{itemize}

  \item[\texttt{robot.capture\_image()}]
  Return the mock image from the dataset without a real robot.
  \begin{itemize}
    \item \textbf{Inputs}: No input required
    \item \textbf{Outputs}:
      \begin{itemize}
        \item \textbf{Raw value}: \texttt{image\_shape} (numpy array or list), \texttt{image} (PIL Image)
        \item \textbf{Text}: Image dimensions and capture status
        \item \textbf{Image}: Captured image from robot camera
        \item \textbf{Variables}: \texttt{captured\_image} (PIL Image)
      \end{itemize}
  \end{itemize}

  \item[\texttt{robot.get\_depth()}]
  Retrieve depth map from the robot's depth sensor.
  \begin{itemize}
    \item \textbf{Inputs}: No input required
    \item \textbf{Outputs}:
      \begin{itemize}
        \item \textbf{Raw value}: \texttt{depth\_map} (numpy array), \texttt{depth\_map\_visualization} (PIL Image), \texttt{focal\_length\_px} (float), \texttt{width} (int), \texttt{height} (int)
        \item \textbf{Text}: Summary of depth data including image dimensions, focal length, and depth statistics
        \item \textbf{Image}: A colorized depth map visualization where closer objects appear cooler (blue/purple) and distant objects appear warmer (red/yellow)
        \item \textbf{Variables}: \texttt{depth\_map} (2D numpy array of depth values in meters), \texttt{foca\_length\_px} (float, estimated focal length in pixels)
      \end{itemize}
  \end{itemize}
  
  \item[\texttt{robot.\\get\_depth\_with\_pointcloud()}]
  Retrieve depth map from robot's depth sensor and generate 3D point cloud.
  \begin{itemize}
    \item \textbf{Inputs}: No input required
    \item \textbf{Outputs}:
      \begin{itemize}
        \item \textbf{Raw value}: \texttt{image} (PIL Image), \texttt{depth\_map} (numpy array), 
        \texttt{point\_cloud} (numpy array),
        \texttt{focal\_length\_px} (float), \texttt{width} (int), \texttt{height} (int).
        \item \textbf{Text}: Summary of depth data and and point cloud generation including image dimensions, focal length, depth statistics, and point cloud size
        \item \textbf{Image}: A colorized depth map visualization where closer objects appear cooler (blue/purple) and distant objects appear warmer (red/yellow)
        \item \textbf{Variables}: \texttt{depth\_map} (2D numpy array of depth values in meters), 
        \texttt{point\_cloud} ($N\times3$ array of 3D points),
        \texttt{foca\_length\_px} (float, estimated focal length in pixels)
      \end{itemize}
  \end{itemize}
  
    \item[\texttt{robot.execute\_grasp(grasp\_pose)}]
  Execute a grasp by moving the robot to the specified pose via a pre-grasp point, and closing the gripper.
  \begin{itemize}
    \item \textbf{Inputs}:
      \begin{itemize}
        \item \texttt{grasp\_pose}: $4\times4$ transformation matrix representing the grasp pose in the robot's camera frame, OpenCV convention
      \end{itemize}
    \item \textbf{Outputs}:
      \begin{itemize}
        \item \textbf{Raw value}: \texttt{success} (boolean), \texttt{execution\_time\_s} (float), \texttt{image} (PIL Image)
        \item \textbf{Text}: Status of the grasp execution
        \item \textbf{Image}: View from robot camera after the grasp is executed
        \item \textbf{Variables}: \texttt{captured\_image} after the grasp is executed
      \end{itemize}
  \end{itemize}
  \item[\texttt{robot.\\place\_object\_at\_2d\_location(point\_2d)}]
  Move the robot to a place it's currently held object based on a 2D normalized image coordinate. The tool will convert to a 3D placement location automatically by shooting a ray.
  \begin{itemize}
    \item \textbf{Inputs}:
      \begin{itemize}
        \item \texttt{point\_2d}: 2D normalized image coordinate where the object should be placed
      \end{itemize}
    \item \textbf{Outputs}:
      \begin{itemize}
        \item \textbf{Raw value}: \texttt{success} (boolean), \texttt{execution\_time\_s} (float), \texttt{image} (PIL Image)
        \item \textbf{Text}: Status of the release operation
        \item \textbf{Image}: View from robot camera after the placement is executed
        \item \textbf{Variables}: \texttt{captured\_image} (PIL Image) after the placement is executed
      \end{itemize}
  \end{itemize}

  \item[\texttt{robot.\\place\_object\_at\_3d\_location(point\_3d)}]
  Move the robot to a 3D placement point and open the gripper to place the object.
  \begin{itemize}
    \item \textbf{Inputs}:
      \begin{itemize}
        \item \texttt{point\_3d}: 3D point in the robot's camera frame (list or numpy array) where the object should be placed
      \end{itemize}
    \item \textbf{Outputs}:
      \begin{itemize}
        \item \textbf{Raw value}: \texttt{success} (boolean), \texttt{execution\_time\_s} (float), \texttt{image} (PIL Image)
        \item \textbf{Text}: Status of the placement operation
        \item \textbf{Image}: View from robot camera after the placement is executed
        \item \textbf{Variables}: \texttt{captured\_image} (PIL Image) after the placement is executed
      \end{itemize}
  \end{itemize}

\end{description}
\FloatBarrier

%% file: tex/appendix/3_method.tex
\section{Additional Method Details}
\label{app-sec:method}

\subsection{Group Relative Policy Optimization}
We employ Group Relative Policy Optimization (GRPO)~\cite{grpo} as our RL training algorithm.
We present the details of GRPO below.

For each input $\mathcal{I}$, in total $N$ rollout procedures are launched asynchronously under the current policy $\pi_{\theta}$. 
Each rollout generates in total $N$ multi-turn rollouts $O_1$, $O_2$, $\dots$, $O_N$. Their rewards are calculated as $r_1$, $r_2$, $\dots$, $r_N$.  
Each \(r_i\) is standardized into a relative advantage \(A_i\) via group computation:
\begin{equation}
A_i = \frac{r_i - \text{mean}(\{r_1, r_2, \ldots, r_N\})}{\text{std}(\{r_1, r_2, \ldots, r_N\})}.
\end{equation}
The policy is then optimized by minimizing the GRPO loss:
\begin{equation}
\begin{aligned}
\mathcal{L}_{\text{GRPO}}(\theta)
&= \mathbb{E}_{i}\Big[
 -\min\!\big(\rho_i A_i,\,
 \text{clip}(\rho_i, 1{-}\epsilon, 1{+}\epsilon)A_i\big) \\
&\qquad +\, \beta\, \text{KL}\!\big(\pi_\theta \,\|\, \pi_{\text{ref}}\big)
\Big],
\end{aligned}
\label{eq:grpo}
\end{equation}
where $\rho_i = \tfrac{\pi_\theta(i)}{\pi_{\text{ref}}(i)}$, and $\pi_{\text{ref}}$ denotes the reference policy model, i.e., the VLM trained after stage 1.  
Here, $\epsilon$ and $\beta$ are tunable hyperparameters controlling the clipping range and KL regularization strength.  
This formulation encourages the policy to increase the probability of high-reward responses while maintaining stability through KL regularization.

\begin{figure}[t]
\centering
\includegraphics[width=0.8\linewidth]{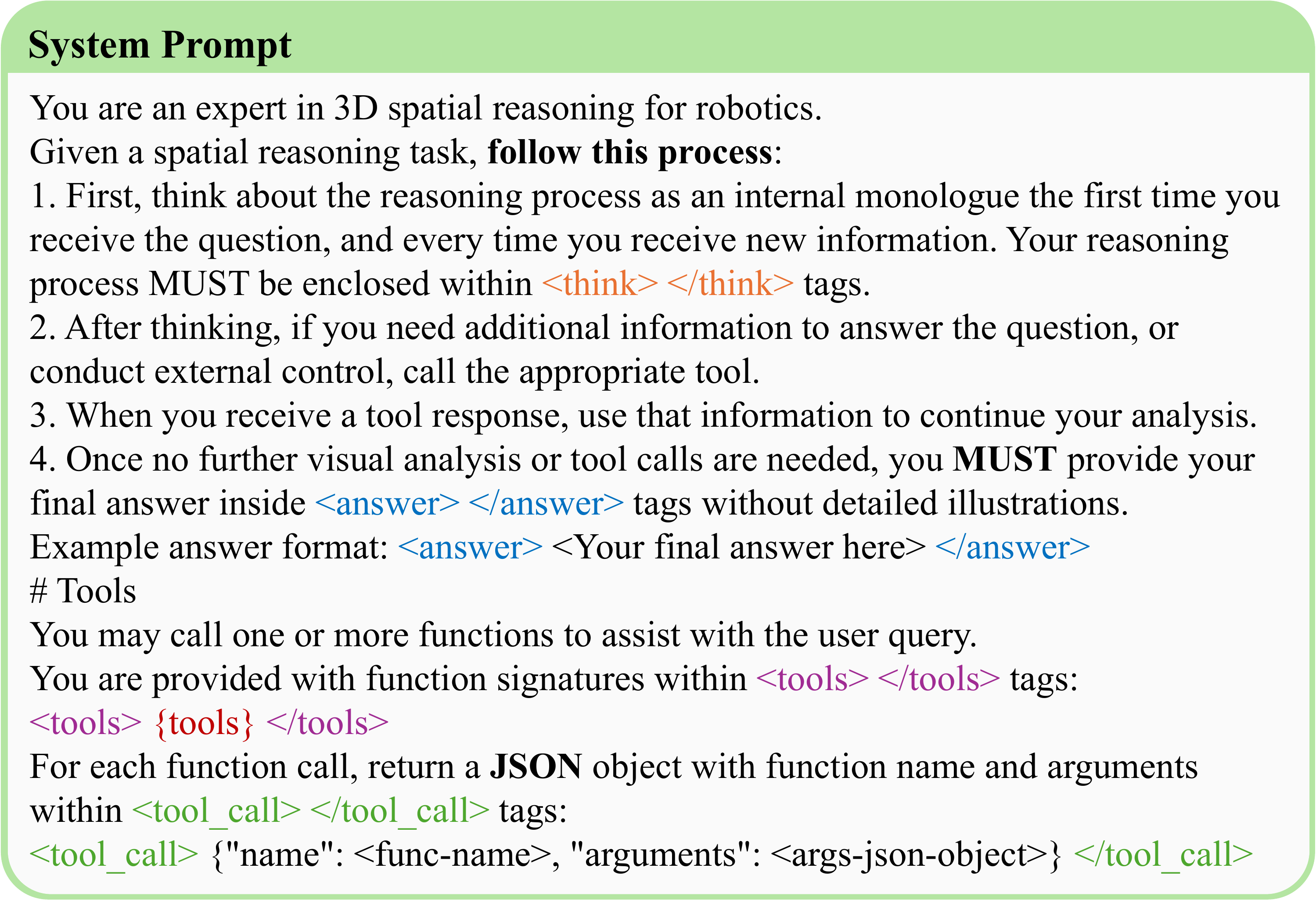}
\caption{\textbf{System prompt.} 
Instructional prompt guiding the model's reasoning, tool-call, and answer process. 
}
\label{fig:sys_prompt}
\end{figure}

\subsection{Alternative Reward Design}
\label{appsubsec:reward}

\paragraph{Other reward for the pointing questions.} Considering that pointing is the first step of solving many spatial reasoning tasks or using other tools, we have experimented with several different rewards for pointing before finalizing the NNDC reward. We show results in Appendix~\ref{appsec:ablation}, emphasizing the importance the reward design for tasks requiring explicit numerical estimation.
\begin{itemize}
\item \textbf{Binary:} 
    \[
    R_{\text{B}} =
    \begin{cases}
    1.0, & \text{if the predicted point lies within 
    }\\& \text{the ground truth convex hull}, \\
    0, & \text{otherwise}.
    \end{cases}
    \]

\item \textbf{Normalized Signed Distance to Hull (NSDH):}
    \[
    R_{\text{NSDH}} =
    \begin{cases}
    0.5 + 0.5\exp(s), & \text{if } s \leq 0, \\
    0.5\exp(-s), & \text{if } s > 0,
    \end{cases}
    \]
    where $s$ is the signed distance from the predicted point to the convex hull boundary (negative inside, positive outside).

    \item \textbf{Normalized Area Change (NAC):}
    Let $\Delta A$ be the change in area after adding the predicted point to the convex hull, and $A_0$ be the original area. Then:
    \[
    R_{\text{NAC}} = \exp\left(-\frac{\Delta A}{A_0}\right).
    \]
\end{itemize}
Similar to NNDC, by default, we also apply clipping with the binary accuracy term to emphasize precision for alternative non-binary rewards. (\textit{e.g.}, $R = \max(R_{\text{NSDH}}, R_{\text{B}})$)

\paragraph{Format score details.}

In addition to task-specific rewards, we explored defining a \emph{format score} to enforce the structural correctness of model outputs (as defined in the system prompt shown in Figure~\ref{fig:sys_prompt}), but did not use it in the final training. 
The format score verifies that every \texttt{<tool\_call>} tag is immediately preceded by a \texttt{<think>} tag, and that the single final \texttt{<answer>} is also directly preceded by a \texttt{<think>}. 
The output must contain exactly one \texttt{<answer>} tag, positioned at the end of the response. 
Additional optional constraints can be applied, such as requiring at least one \texttt{<tool\_call>} in the output. 
Predictions failing to meet these structural requirements receive a format score of zero, while perfectly formatted predictions receive a score of one.

The final reward is computed as a weighted sum of the accuracy-based reward and the format score:
\[
R_{\text{final}} = R_{\text{acc}} + \lambda \, R_{\text{format}},
\]
where $R_{\text{acc}}$ denotes the task-specific accuracy reward, $R_{\text{format}} \in \{0, 1\}$ is the format score (equal to 1.0 if all structural criteria described above are satisfied, and 0 otherwise), and $\lambda \in [0, 1]$ is a scalar weight controlling the influence of the format score. Following prior work, $\lambda$ is often set to $0.3$.

%% file: tex/appendix/4_implementation.tex
\section{Additional Implementation Details}
\label{sec:app-implementation-details}

\subsection{More Training and Compute Details}

\paragraph{Prompts and teaching data.}
For Phase-1 SFT of the base model and Phase-2 IRL, we use the system prompt shown in Figure~\ref{fig:sys_prompt}. To improve the effectiveness of Phase-1 IRL training for the teacher model, we augment this prompt with two tool-use examples: one demonstrating how to solve a spatial relationship problem using the pointing tool, and another illustrating 2D bounding box estimation with the same tool. In addition, when generating teaching data with Claude Sonnet 4.5, we include extra instructional prompts that encourage careful interpretation of tool outputs and better understanding of image coordinate systems.

\paragraph{Synthetic robot data for teaching.}
Due to the substantial latency of robot-in-the-loop training and data collection, we construct grasp and place data using the HOPE dataset and provide it to Claude Sonnet 4.5 together with mock robot tools to generate the robotic portion of the teaching dataset. The interactive learning stages themselves do not use this synthetic robot-tool component; instead, they focus exclusively on spatial reasoning with vision tools.

\paragraph{Answer balancing.}
Another practical consideration is maintaining balanced answer distributions for multiple-choice questions. For example, the original RoboSpatial VQA dataset contains more than 75\% ``no'' answers, which biases the VLM toward predicting ``no'' during both SFT and IRL. We find that rebalancing the data mitigates this issue and improves answer calibration across tasks.

\paragraph{Hyperparameters and training stability.}
The major hyperparameters used during training are summarized in Table~\ref{tab:training_phases}. Across our experiments, the KL coefficient emerges as a particularly important hyperparameter: a relatively small value is needed to encourage sufficient exploration during RL. However, with older versions of Verl, this choice can introduce training instability. In particular, during Phase-1 IRL, we observe an initial reward drop when the KL coefficient is too small. We explored several mitigation strategies, including format rewards, format penalties, alternative KL loss formulations, and related variants, but none fully eliminated this effect. Interestingly, the instability is alleviated when integrating Toolshed with the new agent loop feature introduced in newer versions of Verl. For this reason, we open-source our code based on integration with the latest Verl fork.

\begin{table}[h]
\centering
\caption{
Training configurations for Phase-1 IRL, Phase-1 SFT, and Phase-2 IRL. 
A dash (--) indicates that the setting is not applicable to that phase.
}
\scriptsize
\begin{tabular}{@{}lccc@{}}
\toprule
& \textbf{Phase-1 IRL} & \textbf{Phase-1 SFT} & \textbf{Phase-2 IRL} \\
\midrule

\rowcolor[HTML]{F5F5F5}
\multicolumn{4}{@{}l}{\textit{Data}} \\
Dataset                 & Direct VQA      & Teaching tool-use   & Direct VQA \\
\#Samples               & 4k              & 8k                  & $\approx$8k \\[2pt]

\midrule
\rowcolor[HTML]{F5F5F5}
\multicolumn{4}{@{}l}{\textit{Model}} \\
Trainable Part          & \multicolumn{3}{c}{Language model (LLM) only; vision encoder + projector frozen} \\
\#Tunable Parameters    & \multicolumn{3}{c}{2.55B} \\[2pt]

\midrule
\rowcolor[HTML]{F5F5F5}
\multicolumn{4}{@{}l}{\textit{Training}} \\
Batch Size              & 64    & 8     & 64 \\
Learning Rate           & 1e-6  & 1e-5  & 1e-6 \\
Epoch                   & 5     & 2     & 2 \\
Warmup Ratio            & 0.0   & 0.1   & 0.0 \\
LR Schedule             & NA    & cosine & NA \\
KL Coefficient          & 1e-4  & --    & 1e-4 \\
Entropy Coefficient     & 0.0   & --    & 0.0 \\
Temperature             & 1.0   & --    & 1.0 \\
Max Prompt Length       & 8192  & 8192  & 8192 \\
Max Response Length     & 8192  & 8192  & 8192 \\
Rollout Number          & 5     & --    & 5 \\
\#GPU (VLM)             & 8    & 8     & 8 \\
\#GPU (Tools)           & 8    & --    & 8 \\

\bottomrule
\end{tabular}
\label{tab:training_phases}
\end{table}

\subsection{MACE Metric for Grasp Affordances}
The grasp estimation task requires the model to predict five key points in normalized image coordinates: the grasp center, left finger base, right finger base, left finger tip, and right finger tip. From these points, we define four finger direction vectors: grasp center $\rightarrow$ left finger base, grasp center $\rightarrow$ right finger base, left finger base $\rightarrow$ left finger tip, and right finger base $\rightarrow$ right finger tip.

The Mean Angular Coordinate Error (MACE) metric is defined as follows. Given the predicted and ground-truth grasp centers $\hat{c}$ and $c$, and the set of four corresponding finger direction vectors $\{\hat{r}_k\}_{k=1}^4$ and $\{r_k\}_{k=1}^4$, we define:
\begin{equation}
\text{MACE}
= 1 - \frac{1}{2} \left(
\frac{\lVert \hat{c} - c \rVert_2}{w} + 
\frac{1}{4} \sum_{k=1}^{4} \frac{1 - \cos(\hat{r}_k, r_k)}{2}
\right),
\label{eq:mace}
\end{equation}
where $w$ denotes the gripper width used for spatial normalization, and $\cos(\hat{r}_k, r_k) = \frac{\hat{r}_k \cdot r_k}{\lVert \hat{r}_k \rVert \lVert r_k \rVert}$ represents the cosine similarity between the predicted and ground-truth orientations of the $k$-th finger-related vector.

\subsection{Robot Manipulation Setup}

\begin{table}[t]
\centering
\small
\begin{tabular}{p{6.6cm}ccc}
\toprule
\textbf{Task} & \textbf{Ours} & \textbf{Claude} & \textbf{GPT} \\
\midrule
\multicolumn{4}{l}{\textbf{Pick}} \\
Pick up the dark blue object & \gmark & \gmark & \gmark \\
Pick up the soft toy & \rmark & \rmark & \rmark \\
Pick up the solid toy & \gmark & \gmark & \rmark \\
Pick up the tall cylindrical tennis ball container & \gmark & \gmark & \gmark \\
Pick up the coconut water & \gmark & \gmark & \gmark \\
Pick up the plastic bottle & \gmark & \gmark & \gmark \\
Pick up the red box & \gmark & \gmark & \gmark \\
\midrule
\multicolumn{4}{l}{\textbf{Relational Pick}} \\
Pick up the far coconut water & \gmark & \rmark & \rmark \\
Pick up the coconut water that is closer to the camera & \gmark & \gmark & \rmark \\
Pick up the left pineapple juice can & \gmark & \rmark & \rmark \\
Pick up the right pineapple juice can & \gmark & \gmark & \rmark \\
Pick up the further purple drink & \rmark & \gmark & \gmark \\
Pick up the near purple bottle & \gmark & \rmark & \gmark \\
\midrule
\multicolumn{4}{l}{\textbf{Pick \& Place}} \\
Pick up the hot dog and place it in the black bin & 0 & 2 & 2 \\
Pick up the tall cylindrical container and place it in the transparent bin & 2 & 2 & 2 \\
Pick up the leftmost condiment and place it in the transparent bin & 2 & 2 & 2 \\
Pick up the cinnamon and place it in front of the rice box & 2 & 2 & 0 \\
Pick up the rice box and place it next to the hot sauce & 2 & 1 & 0 \\
Pick up the plushie and place it left of the coconut water & 2 & 1 & 1 \\
Pick up the pony and place it left of the two plushies & 2 & 1 & 2 \\
\bottomrule
\end{tabular}
\caption{Per-task breakdown of the real-world manipulation results, comparing \textbf{Ours} (\modelname{}), Claude Sonnet 4.5 and GPT-5.}
\label{app:tab-robot-results}
\end{table}

\paragraph{Robot System}
We conduct robot experiments on a Kinova Jaco
arm equipped with the CuRobo~\cite{curobo} motion planner and a ZED2 RGB-D camera. We expose the robot as a tool and make it available to the VLM. The tool has functions: \texttt{capture\_image}  retrieves the current RGB image from the camera, \texttt{get\_depth} and \texttt{get\_depth\_with\_pointcloud} retrieve the current depth image, optionally with a pointcloud in the robot frame, \texttt{execute\_grasp} moves the end-effector to a specified grasp pose via a pre-grasp point and closes the gripper, \texttt{place\_object\_at\_2d\_location} and \texttt{place\_object\_at\_3d\_location} offer two ways to parameterize a place operation that moves the robot hand holding an object over a location in the scene and opens the gripper.
All motions are executed with the motion planner. 

\paragraph{Robot Experiments Tasks and Results}
We design a suite of tasks across three categories. \emph{Pick}, \emph{Relational Pick}, and \emph{Pick \& Place}. 
We score both \emph{Pick} tasks based on the success rate, and \emph{Pick \& Place} based on partial success rate, awarding 1 point each for a correct pick and place operation.

The full results at individual task level are in~\Cref{app:tab-robot-results}, omitting methods that fail to achieve any points.
In \emph{Pick up the soft toy} task, all models failed due to a common failure in pointing tool not being able to differentiate the soft toy from a rigid toy.
In \emph{Relational Pick} and \emph{Pick \& Place} tasks, \modelname{} shows superior ability than Claude and GPT-5 in correctly using the pointing tool to resolve spatial relations, reflecting an understanding of its strengths and limitations likely attributable to the interactive training with the tool.

\paragraph{Additional Details on Robot SFT Data Collection}
In order to collect SFT data of from the Universal Teacher (Claude Sonnet 4.5) on using the robot tool, we design a ``mock robot''  that has the same API as the robot tool, but it always simulates successful actions provided the API was called with valid arguments. This allows collecting a small number ($\sim$500) examples of valid robot API calls without requiring the physical robot in the loop and ensuring that our robot is unseen during training.

\subsection{Details of Non-interactive RL Baseline}
We present the detailed description of the Tool NIRL baseline referenced in the ablation section of the main paper, as space limitations prevented us from including all details there. 

We follow the conventional tool-learning setup used in LLMs \cite{zhang2025nemotronresearchtooln1exploringtoolusinglanguage} to perform reinforcement learning of tool usage \emph{without} executing tools during training. The core idea is to compute the reward solely from the correctness of the predicted tool name and its arguments, which requires access to ground-truth tool call traces for supervision. After obtaining tool-augmented reasoning traces from Claude, each multi-turn trace with $T$ turns is decomposed into $T$ single-turn training instances: the $i^{\text{th}}$ instance contains the conversation history up to turn $i$ as input, and the corresponding ground-truth output for turn $i$ as the target.

During training, for tool-call turns, we adopt the binary reward used in \cite{zhang2025nemotronresearchtooln1exploringtoolusinglanguage}. A reward of $1.0$ is given only when both the tool call format and the tool call content match the ground truth:
\begin{equation}
r =
\begin{cases}
1, & \text{if } \texttt{FormatCorrect} \wedge \texttt{ToolCallMatch} \\
0, & \text{otherwise},
\end{cases}
\end{equation}
where \texttt{FormatCorrect} verifies that the model output is wrapped in the required tags, and \texttt{ToolCallMatch} checks that both the tool name and its arguments exactly match the ground-truth tool call. For final-answer turns (i.e., non-tool turns), we reuse the same task-specific normalized rewards introduced in this paper.

%% file: tex/appendix/5_results.tex
\section{Additional Experimental Results}
\label{sec:app-exp}

\begin{table}[t]
\centering
\scriptsize
\caption{Benefit of scaling tool instances with \systemname{} under contention. We measure 8 simultaneous RoboRefer tool calls. Compared with a naive HTTP-based deployment using a single instance, \systemname{} with 3 instances substantially reduces end-to-end latency.}
\begin{tabular}{@{}p{4.2cm}cc@{}}
\toprule
\textbf{Metric} & \textbf{Naive HTTP (1 instance)} & \textbf{\systemname{} (3 instances)} \\
\midrule
Wall-clock Time $\downarrow$ & $8.5 \pm 0.3$ s & $\mathbf{2.7 \pm 0.1}$ s \\
Speedup $\uparrow$ & baseline & $\mathbf{3.2\times}$ \\
\bottomrule
\end{tabular}
\label{tab:toolshed_scaling_roborefer}
\end{table}

\begin{table}[t]
\centering
\scriptsize
\caption{Pipeline execution latency for answering ``Is bok choy or clock closer?'' using 2$\times$ RoboRefer, 2$\times$ SAM, 1$\times$ depth estimation, and 2$\times$ \texttt{index\_at} tools. \systemname{} improves both wall-clock latency and throughput, with further gains when scaling to 3 tool instances.}
\begin{tabular}{@{}p{4.0cm}ccc@{}}
\toprule
\textbf{Metric} & \textbf{HTTP (1 inst)} & \textbf{\systemname{} (1 inst)} & \textbf{\systemname{} (3 inst)} \\
\midrule
Wall-clock Time $\downarrow$ & $20.23 \pm 0.42$ s & $15.13 \pm 0.08$ s & $\mathbf{10.56 \pm 0.01}$ s \\
Throughput $\uparrow$ & 0.40 pipe/sec & 0.53 pipe/sec & $\mathbf{0.76}$ pipe/sec \\
Per-Pipeline Time $\downarrow$ & 15,678 ms & 11,998 ms & $\mathbf{7,931}$ ms \\
\bottomrule
\end{tabular}
\label{tab:toolshed_pipeline_latency}
\end{table}

\begin{table}[t]
\centering
\scriptsize
\caption{I/O time breakdown (ms) by tool when no queueing is required. \systemname{} is especially advantageous for tools with larger inputs/outputs, such as SAM and depth estimation, while maintaining low overhead across all steps.}
\begin{tabular}{@{}p{4.2cm}cc@{}}
\toprule
\textbf{Step} & \textbf{HTTP I/O} & \textbf{\systemname{} I/O} \\
\midrule
\texttt{roborefer\_bok\_choy} & $6 \pm 0$ & $10 \pm 1$ \\
\texttt{roborefer\_clock} & $7 \pm 0$ & $10 \pm 0$ \\
\texttt{sam2\_bok\_choy} & $18 \pm 2$ & $11 \pm 0$ \\
\texttt{sam2\_clock} & $18 \pm 1$ & $11 \pm 0$ \\
\texttt{depth\_estimation} & $45 \pm 6$ & $11 \pm 0$ \\
\texttt{index\_*} & $0$ & $9 \pm 0$ \\
\bottomrule
\end{tabular}
\label{tab:toolshed_io_breakdown}
\end{table}

\subsection{Efficiency of Toolshed system.}
A naive integration of computer vision tools in Verl is not feasible, since Verl instantiates tools when calling them, and parameter loading can take 5-10 minutes. The most compatible integration might be serving each tool with an HTTP server. We ran an experiment executing a pipeline of 2x pointing, 2x SAM, 2x depth, and 2x indexing tools, with overall results summarized in \Cref{tab:toolshed_scaling_roborefer}, \Cref{tab:toolshed_pipeline_latency}, and \Cref{tab:toolshed_io_breakdown}. The latency of Toolshed is better for large inputs/outputs, like we encounter for computer vision tools. Toolshed also provides more efficient queuing for tools when 8 callers are contending for 1 or 3 available tools. Other than performance, toolshed has additional advantages like seamless development, allowing to write pythonic tools (inputs/outputs can be any python objects, including gpu tensors), scaling to multiple tool instances with a single config change, and python environment isolation.

\input{Tables/interactive_RL}

\subsection{A Closer Look at Tool IRL Alone.}

While direct IRL over diverse tools poses challenges due to the vast action space, we demonstrate its effectiveness within a constrained setup using the RoboSpatial dataset and pointing tools. 
As shown in \Cref{tab:qwen3b_spatial_subset}, this approach substantially improves spatial reasoning compared with both direct tool-free SFT and vanilla tool-free GRPO baselines (e.g., classic reasoning models like R1~\cite{deepseekr1}), 
as well as other inference approaches \cite{CoT}. 
On RoboSpatial, the IRL with Tools model achieves 72.3\% overall accuracy, outperforming SFT and vanilla GRPO. 
Notably, IRL with Tools is the only method that generalizes to unseen tasks: achieving 34.3\% on RefSpatial, whereas other fine-tuning strategies yield zero accuracy. 
These results show that interactive tool use during RL enables the model to internalize transferable geometric reasoning skills beyond the training domain. 
Even without fine-tuning, connecting \systemname{} to the pretrained model yields measurable improvements on RefSpatial, highlighting the intrinsic generalization benefit of tool-augmented spatial reasoning.

\subsection{Scaling to New Tools with DIRL}

DIRL supports learning a more diverse and adaptive set of tools than prior work, while also enabling new tools to be added without fully retraining the entire system. Even in the worst case, only partial retraining is needed. In particular, when introducing a new tool, the interactive RL stage for the foundational pointing tool---which accounts for roughly 60\% of the total training compute---can always be reused. As a result, the worst-case cost of incorporating a new tool is only about 40\% of the compute required for full DIRL training. See results in \Cref{tab:dirl_new_tool_scaling}.

We further demonstrate there a more efficient strategy for new-tool adaptation. Starting from a DIRL model trained without the grasp prediction tool, we extend it using only supervised fine-tuning (SFT) and the final IRL stage on grasp data. This lightweight recipe requires only about 10\% of the full training compute while achieving comparable performance. As a representative case study, we evaluate the addition of a grasp prediction tool. A broader investigation of continual learning with newly introduced tools and tasks remains an important direction for future work.

\begin{table}[t]
\centering
\scriptsize
\caption{Adding a new tool to DIRL using grasp prediction as a representative example. Even in the worst case, DIRL only requires partial retraining, since the interactive RL stage for the foundational pointing tool can be reused. We also show an efficient adaptation recipe that uses only SFT and the final IRL stage on grasp data.}
\begin{tabular}{@{}lccc@{}}
\toprule
\textbf{Setting} & \textbf{RoboSpatial} & \textbf{Grasp-SR} & \textbf{Compute (\% of full DIRL)} \\
\midrule
Worst case & 70\% & 50\% & $\sim$40\% \\
Efficient example & 62\% & 50\% & $\sim$10\% \\
\bottomrule
\end{tabular}
\label{tab:dirl_new_tool_scaling}
\end{table}

\begin{figure*}[t]
    \centering
    \includegraphics[width=\linewidth]{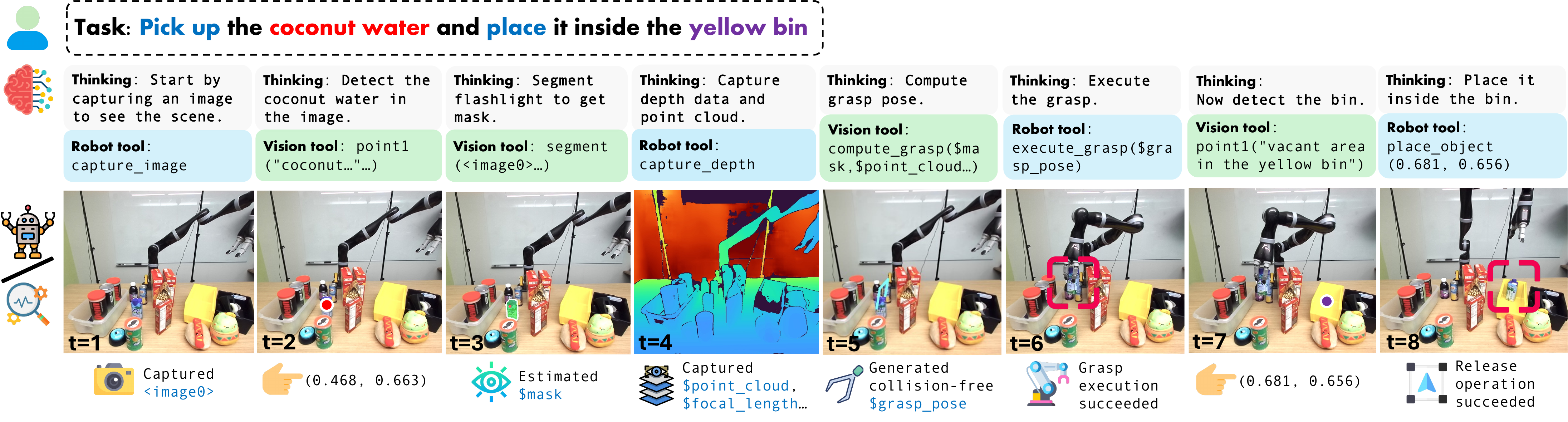}
    \caption{A hard real-world robot manipulation example with \modelname{}. The model successfully identifies the target object and completes the manipulation task in a cluttered and visually complex scene
    }
    \label{fig:robot_exp2}
\end{figure*}

\begin{figure*}[t]
    \centering
    \includegraphics[width=\linewidth]{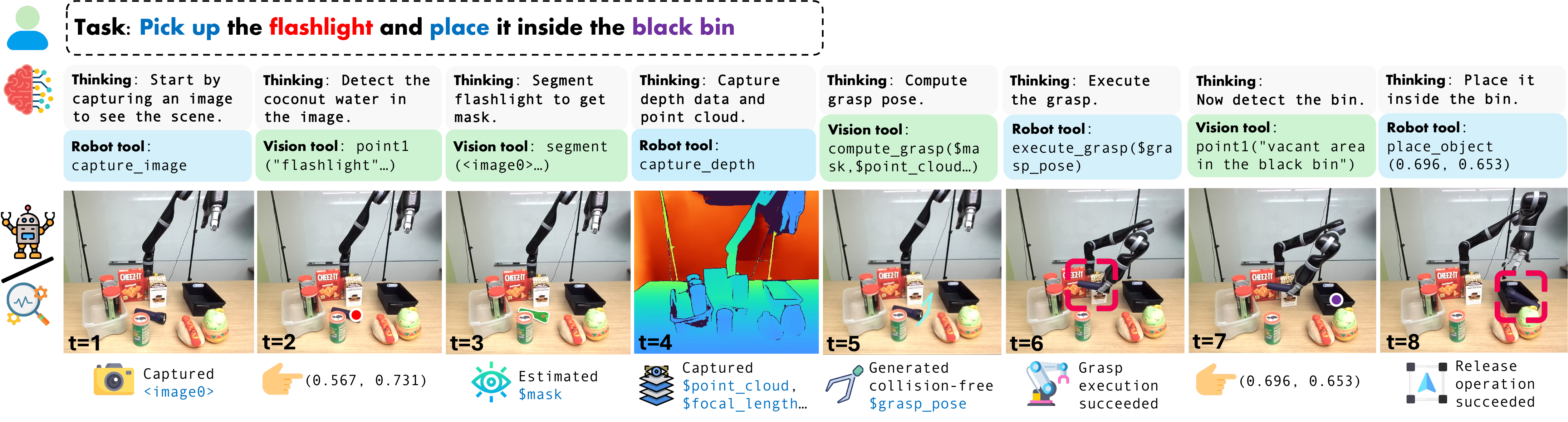}
    \caption{A failure case in real-world robot manipulation with \modelname{}. The model localizes a valid vacant area but selects a point too close to the boundary, resulting in a failed placement of the flashlight on the box boundary.}
    \label{fig:robot_exp3}
\end{figure*}

\begin{figure*}[t]
    \centering
    \includegraphics[width=\linewidth]{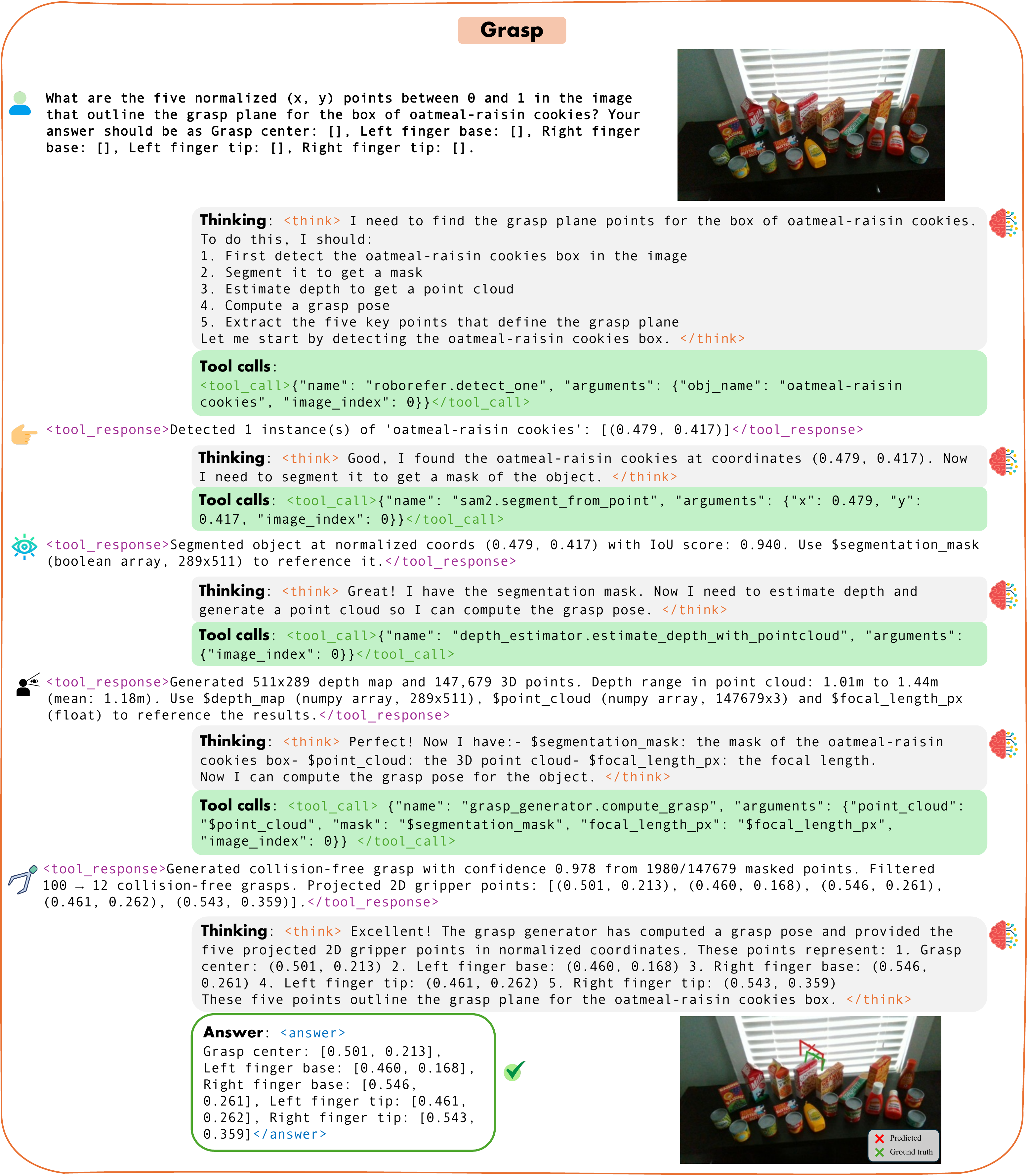}
    \caption{A detailed example of tool-augmented reasoning of a grasp estimation question.}
    \label{fig:grasp_detail}
\end{figure*}

\begin{figure*}[t]
    \centering
    \includegraphics[width=\linewidth]{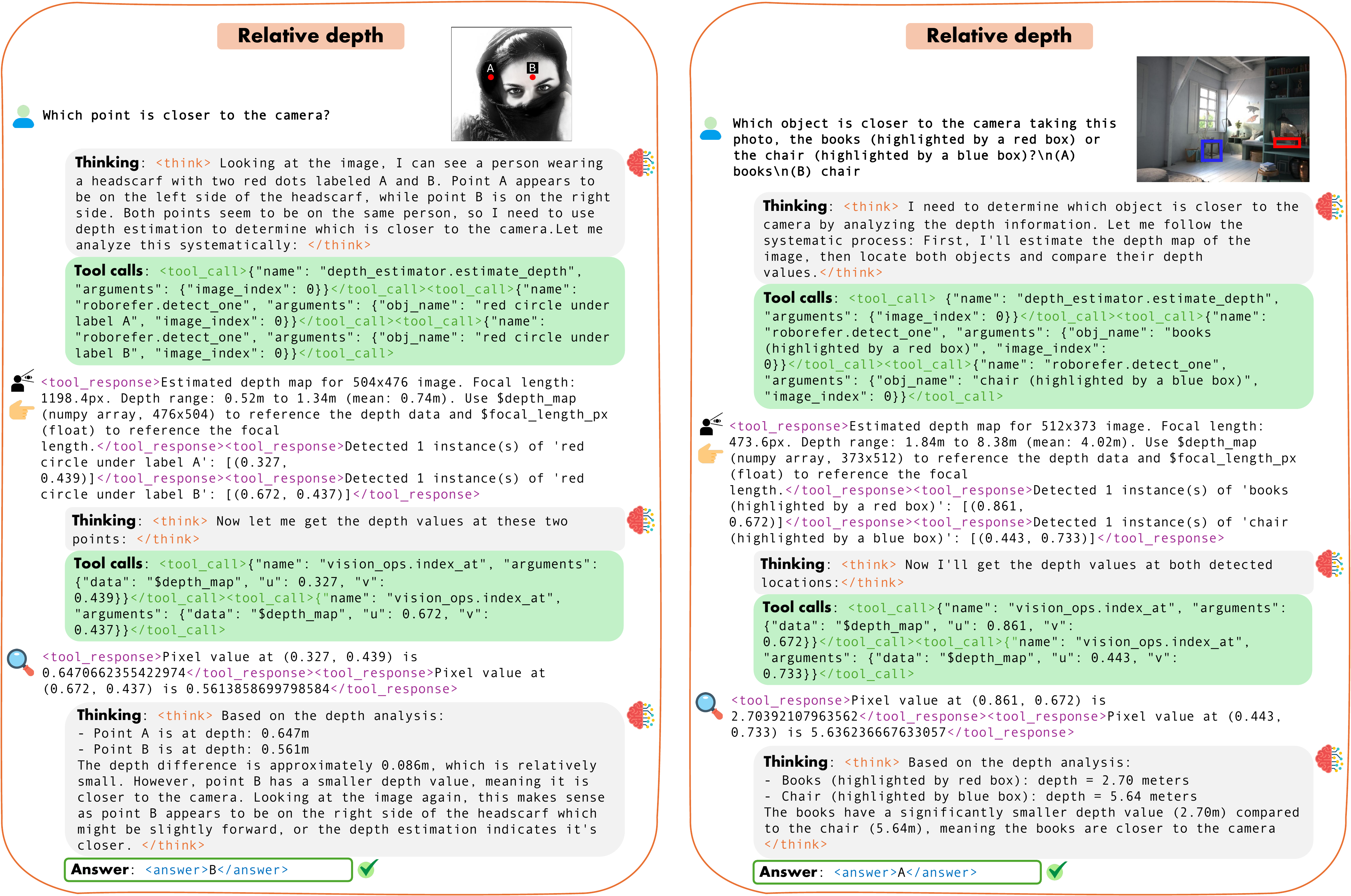}
    \caption{Detailed examples of tool-augmented reasoning of relative depth questions.}
    \label{fig:depth_detail_combine}
\end{figure*}

\subsection{Other Ablations}
\label{appsec:ablation}

\paragraph{Direct IRL on all tasks with all tools.}

As mentioned in the main paper, directly reinforcement learning with all tools on all tasks result in a large search space and is hard to learn effectively. We provide qualitative performance in Table~\ref{tab:s2rlonly}, supporting this argument.

\begin{table}[h]
\caption{Direct IRL on all tasks (\textit{Direct IRL All.}) with all tools compared with our method.}
\centering
\scriptsize
\setlength{\tabcolsep}{4pt}
\renewcommand{\arraystretch}{1.15}

\begin{tabular}{@{}lccc cccc@{}}
\toprule
\textbf{Variant} & 
\textbf{IRL-T} & 
\textbf{Univ-T} &
\textbf{S2-IRL} &
\textbf{RoboSpatial} & 
\textbf{RefSpatial} & 
\textbf{Pose} & 
\textbf{Mean} \\
\midrule

SpaceTools (Ours) & 
\gmark & \gmark & \gmark &
70.00 & 53.07 & 34.37 & 52.48 \\

Direct IRL All. & 
\rmark & \rmark & \gmark &
52.86 & 3.25 & 3.26 & 19.79 \\

\bottomrule
\end{tabular}
\label{tab:s2rlonly}
\end{table}

\paragraph{Reward and prompt for IRL.}
Due to computational constraints that prevent running full-scale IRL ablations, we evaluate different pointing rewards (\textit{e.g.}, NSDH, NAC, Binary) and prompt design choices introduced in Appendix~\ref{appsubsec:reward} on a subset of 1k vacant-space localization questions from RoboSpatial using the Molmo pointing tool. The results in Table~\ref{tab:ablation_reward} show that NNDC without an additional format reward yields the most stable and reliable learning behavior. Accordingly, we adopt NNDC (without a format reward) for all subsequent training stages. These experiments also highlight the importance of normalizing rewards to the 
$[0, 1]$ range, a practice we apply consistently across all tasks. More broadly, this study underscores the richness of the reward-design space for spatial reasoning tasks, especially those requiring explicit numerical estimation.

\begin{table}[h]
\centering
\scriptsize
\setlength{\tabcolsep}{4.5pt}
\renewcommand{\arraystretch}{1.15}
\caption{Ablation on reward and prompt design for the pointing task as introduced in Appendix~\ref{appsubsec:reward}. \textit{Norm.} indicates whether normalization to range $[0,1]$ is applied to the reward function. \textit{Clip.} indicates whether binary clipping is applied. \textit{Format} indicates whether the format reward is applied. \textit{Example in Prompt} indicates whether two tool-use examples are added in the prompt.
Checkmarks indicate which components are included for each variant.}
\begin{tabular}{@{}lccccc@{}}
\toprule
\textbf{Reward Variant} & \textbf{Norm.} & \textbf{Clip.} & \textbf{Format} & \textbf{Example in Prompt} & \textbf{Acc.} \\
\midrule

\textbf{NNDC (Ours)}             
& \gmark & \gmark & \rmark & \gmark &  \textbf{35.25} \\

\hspace{0.5em} w/o Clip.      
& \gmark & \rmark & \rmark & \gmark & 14.8 \\

\hspace{0.5em} w/o Norm.  
& \rmark & \gmark & \rmark & \gmark & 0.00 \\

\hspace{0.5em} w Format.      
& \gmark & \gmark & \gmark & \gmark & 33.61 \\

\hspace{0.5em} w/o Example.      
& \gmark & \gmark & \rmark & \rmark & 17.21 \\
\midrule

\textbf{NSDH}             
& \gmark & \gmark & \rmark & \gmark &  21.31 \\

\hspace{0.5em} w/o Clip.      
& \gmark & \rmark & \rmark & \gmark & 22.31 \\

\hspace{0.5em} w/o Norm.  
& \rmark & \gmark & \rmark & \gmark & 0.00 \\
\midrule

\textbf{NAC}             
& \gmark & \gmark & \rmark & \gmark &  22.95 \\

\hspace{0.5em} w/o Clip.      
& \gmark & \rmark & \rmark & \gmark & 22.95 \\

\hspace{0.5em} w/o Norm.  
& \rmark & \gmark & \rmark & \gmark & 0.00 \\
\midrule

\textbf{Binary}                    
& \gmark & \gmark & \rmark & \gmark & 15.57 \\

\bottomrule
\end{tabular}
\label{tab:ablation_reward}
\end{table}

\paragraph{Dataset size and type for IRL.}
We conduct preliminary experiments on how dataset size and data-type composition affect IRL performance using the RoboSpatial dataset and the Roborefer pointing tool. RoboSpatial contains four data types: configuration, compatibility, grounding, and vacant-space localization. Configuration and compatibility are binary yes/no questions, grounding requires predicting 2D bounding boxes, and vacant-space localization involves predicting a free-space location. We vary the mixture of these four types and evaluate performance on RoboSpatial-Home, with results summarized in Table~\ref{tab:data_composition}. Notably, for configuration and compatibility, we ensure a balanced distribution of yes/no answers, as discussed in Appendix~\ref{sec:app-implementation-details}. Interestingly, although grounding data are not present in RoboSpatial-Home, including grounding during training improves performance on the other tasks. In contrast, increasing the overall dataset size beyond a moderate scale yields limited gains, suggesting that data diversity and label balance contribute more to IRL effectiveness than raw quantity alone.

\begin{table}[h]
\centering
\scriptsize
\setlength{\tabcolsep}{6pt}
\renewcommand{\arraystretch}{1.15}
\caption{
Evaluation on RoboSpatial-Home using models trained with Tool IRL under different data compositions drawn from the four RoboSpatial data types. 
\textit{Config.} refers to configuration data, \textit{Compat.} to compatibility data, \textit{Ground.} to grounding (2D bounding box) data, and \textit{Vacant} to vacant-space localization data. 
Each entry in the middle columns indicates the number of samples included for that data category. 
\textit{Overall Acc.} reports the final accuracy on RoboSpatial-Home.
}
\begin{tabular}{@{}lccccc@{}}
\toprule
\textbf{Variant} &
\textbf{Config.} &
\textbf{Compat.} &
\textbf{Ground.} &
\textbf{Vacant} &
\textbf{Overall Acc.} \\
\midrule

All-v1 & 0.5k & 0.5k & 0.5k & 0.5k & 69.70 \\
 
All-v2 & 1.0k & 1.0k  & 1.0k & 1.0k & 69.70 \\

All-v3 & 2.0k & 2.0k  & 1.0k & 1.0k & 69.10 \\

w/o Ground. & 2.0k & 2.0k & 0.0k & 2.0k & 56.90 \\

\bottomrule
\end{tabular}
\label{tab:data_composition}
\end{table}

\subsection{Additional Demonstrations}

\paragraph{Visualizations of success cases.}

Apart from the cases where the grasp tool fails to find a collision-free pose or the pointing tool fails to localize points accurately for relative depth questions, we also present detailed examples of successful tool executions containing all special format-related tokens in Figure~\ref{fig:grasp_detail}. In this example, the grasp tool successfully finds an accurate pose in a cluttered scene. Figure~\ref{fig:depth_detail_combine} provides examples for relative depth questions. One example shows the tool accurately predicts the required point locations, enabling the VLM to produce the correct answer. The other example shows multiple chairs are present in the image, and the VLM must identify the specific one highlighted by the provided bounding box. Moreover, as illustrated in Figure~\ref{fig:robot_exp2}, \modelname{} can reliably identify target objects and execute the required manipulation steps even in cluttered, visually complex real-world environments.

\begin{figure}[h]
    \centering
    \includegraphics[width=.6\linewidth]{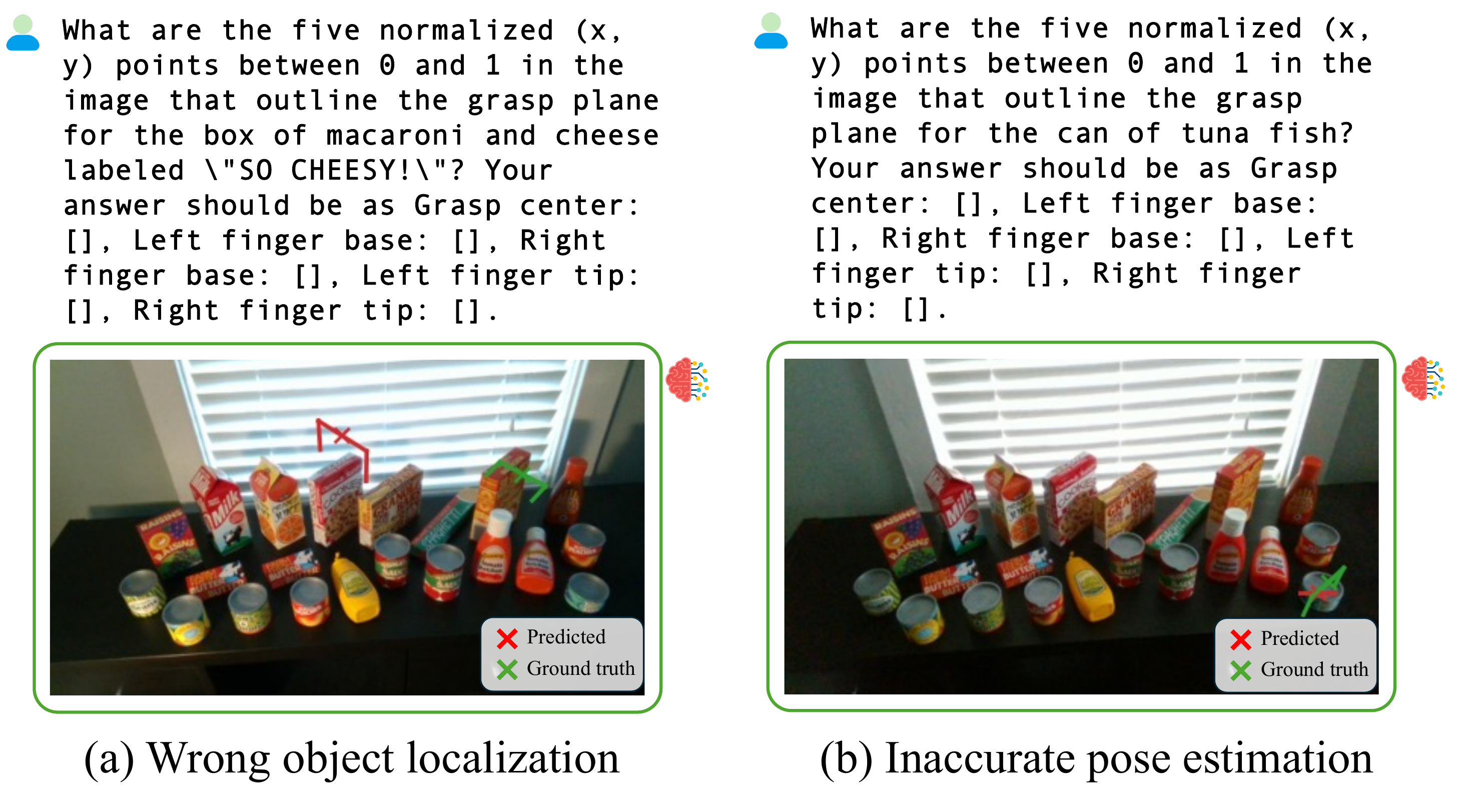}
    \caption{Failure cases for the grasp estimation task. Intermediate tool-augmented reasoning steps are omitted for clarity.}
    \label{fig:grasp_fail}
\end{figure}

\paragraph{Visualizations of failure cases.}
Although the model performs well across many scenarios, it is not universally reliable. We therefore examine representative failure cases to highlight remaining challenges and motivate future exploration. Grasp estimation currently yields the lowest accuracy, largely due to difficulties in detecting the target object in cluttered scenes and in predicting accurate yet collision-free grasp poses, as illustrated in Figure~\ref{fig:grasp_fail}. Object detection in complex, visually crowded environments remains a major bottleneck for both grasp and pose estimation, leaving substantial room for improvement. These issues point not only to opportunities for improving individual tools but also to the potential of modular enhancements to spatial reasoning by strengthening specific tool components instead of retraining or finetuning the VLM. 

Another failure example of real-world robot manipulation is shown in Figure~\ref{fig:robot_exp3}. This failure case highlights a subtle but important challenge in real-world manipulation. Although the model correctly identifies a vacant region, it selects a point near the boundary, and thus, the robot finally performs the placement on the boundary. Such cases underscore the need for highly precise geometric reasoning and tighter coupling between 2D point predictions and physical feasibility on the robot. They also reveal a current limitation of the model's tool coordination and point-selection strategy, suggesting promising directions for future improvements, including integrating real or simulated robot feedback into the training process.

\begin{table}[t]
\centering
\scriptsize
\caption{Representative failure breakdown by primary error source. We classify each failed case according to the most direct cause leading to the incorrect outcome.}
\begin{tabular}{@{}lccc@{}}
\toprule
\textbf{Setting} & \textbf{Tool Error} & \textbf{Planning Error} & \textbf{Reasoning Error} \\
\midrule
Robot Manipulation & 2 & 2 & 0 \\
Grasp Benchmark & 23 & 0 & 7 \\
\bottomrule
\end{tabular}
\label{tab:failure_breakdown}
\end{table}

\paragraph{Quantitative failure analysis.}
We further report representative failure statistics on the grasp benchmark and real-world robot manipulation in \Cref{tab:failure_breakdown}. Specifically, we analyze 30 failures out of 60 grasp benchmark trials and 4 failures out of 27 robot manipulation trials. Since execution traces are often intertwined---for example, a tool failure may trigger tool switching or fallback reasoning---individual failures cannot always be uniquely attributed to a single source. We therefore categorize each failed case by the most direct cause that led to the final incorrect outcome.

%% file: Tables/interactive_RL.tex
\begin{table}[t]
\centering
\scriptsize
\caption{Comparison of Qwen2.5-VL-3B, its inference variants, and fine-tuned models on RoboSpatial and RefSpatial. 
\textit{Direct Inference} refers to answering the question without intermediate reasoning or tool use. 
\textit{CoT} denotes chain-of-thought inference. 
\textit{+Toolshed} indicates tool-augmented inference without any additional training. 
Among all variants, Tool IRL achieves the highest overall accuracy on RoboSpatial and the strongest generalization to RefSpatial.
}
\begin{tabular}{@{}p{3.2cm}cccc@{}}
\toprule
\multirow{2}{*}{\textbf{Model}} 
& \multicolumn{3}{c}{\textbf{RoboSpatial}}  
& \multirow{2}{*}{\textbf{RefSpatial}} \\ 
\cmidrule(lr){2-4}
 & \textbf{VQA} & \textbf{Vacant} & \textbf{Overall} &  \\ 
\midrule
\rowcolor[HTML]{F5F5F5}
\multicolumn{5}{c}{\textit{Inference Baseline (no fine-tuning)}} \\
Direct Inference & 53.07 & 0.00 & 35.71 & 0.00 \\
CoT & 66.67 & 0.00  &  43.71 &  0.00 \\
+\systemname{} & 47.37 & 9.02  & 34.00  & 17.69  \\[2pt]
\midrule
\rowcolor[HTML]{F5F5F5}
\multicolumn{5}{c}{\textit{Fine-tuned on RoboSpatial}} \\
Tool-free SFT & 75.88 & 13.11 & 54.00 & 0.00 \\
Tool-free RL & 72.37 & 20.49 & 54.28 & 0.00 \\
\textbf{Tool IRL} & 77.64 & 62.30 & 72.30 & 34.30 \\
\bottomrule
\end{tabular}
\label{tab:qwen3b_spatial_subset}
\end{table}